\documentclass[10pt,twocolumn,letterpaper]{article}

\usepackage{iccv}
\usepackage{times}
\usepackage{epsfig}
\usepackage{graphicx}
\usepackage{amsmath}
\usepackage{amssymb}
\usepackage{comment}
\usepackage{booktabs}       
\usepackage{multirow}

\usepackage{url}

\usepackage[pagebackref=true,breaklinks=true,letterpaper=true,colorlinks,bookmarks=false]{hyperref}

\iccvfinalcopy 

\def\iccvPaperID{4943} 

\ificcvfinal\fi
\begin{document}

\title{Teacher Guided Architecture Search}

\author{Pouya Bashivan\\
Department of Brain and Cognitive Sciences\\
McGovern Institute for Brain Research\\
MIT \\
{\tt\small bashivan@mit.edu}
\and
Mark Tensen\\
University of Amsterdam\\
{\tt\small mark.tensen@student.uva.nl}
\and
James J DiCarlo\\
Department of Brain and Cognitive Sciences and\\
McGovern Institute for Brain Research\\
MIT \\
{\tt\small dicarlo@mit.edu}
}

\maketitle

\begin{abstract}
   Much of the recent improvement in neural networks for computer vision has resulted from discovery of new networks architectures. Most prior work has used the performance of candidate models following limited training to automatically guide the search in a feasible way. Could further gains in computational efficiency be achieved by guiding the search via measurements of a high performing network with unknown detailed architecture (e.g. the primate visual system)? As one step toward this goal, we use representational similarity analysis to evaluate the similarity of internal activations of candidate networks with those of a (fixed, high performing) \textit{teacher} network. We show that adopting this evaluation metric could produce up to an order of magnitude in search efficiency over performance-guided methods. 
   Our approach finds a convolutional cell structure with similar performance as was previously found using other methods but at a total computational cost that is \textit{two orders of magnitude} lower than Neural Architecture Search (NAS) and more than \textit{four times} lower than progressive neural architecture search (PNAS). 
   We further show that measurements from only $\sim$300 neurons from primate visual system provides enough signal to find a network with an Imagenet top-1 error that is significantly lower than that achieved by performance-guided architecture search alone. These results suggest that representational matching can be used to accelerate network architecture search in cases where one has access to some or all of the internal representations of a \textit{teacher} network of interest, such as the brain's sensory processing networks. 
\end{abstract}

\section{Introduction}
The accuracy of deep convolutional neural networks (CNNs) for visual categorization has advanced substantially from 2012 levels (AlexNet \cite{Krizhevsky2012}) to current state-of-the-art CNNs like ResNet \cite{He2015}, Inception \cite{Szegedy2014}, DenseNet \cite{Huang2016}. This progress is mostly due to discovery of new network architectures. Yet, even the space of feedforward neural network architectures is essentially infinite and given this complexity, the design of better architectures remains a challenging and time consuming task.

Many approaches have been proposed in recent years to automate the discovery of neural network architectures, including random search \cite{pinto2009high}, reinforcement learning \cite{Zoph2017a, Zoph2017b}, evolution \cite{Real2016, Real2018}, and sequential model based optimization (SMBO) \cite{Liu2017, Bergstra2012}. These methods operate by iteratively sampling from the hyperparameter space, training the corresponding architecture, evaluating it on a validation set, and using the search history of those scores to guide further architecture sampling.
But even with recent improvements in search efficiency, the total cost of architecture search is still outside the reach of many groups and thus impedes the research in this area (e.g. some of the recent work in this area has spent more than 20k GPU-hours for each search experiment \cite{Real2018, Zoph2017a}).

What drives the total computational cost of running a search?  For current architectural search procedures (above), the parameters of each sampled architecture must be trained before its performance can be evaluated and the amount of such training turns out to be a key driver in the total computational cost.  Thus, to reduce that total cost, each architecture is typically only partially trained to a \textit{premature} state and its \textit{premature} performance is used as a proxy of its \textit{mature} performance (i.e. the performance it would have achieved if was actually fully trained).  

Because the search goal is high \textit{mature} performance in a task of interest, the most natural choice of an architecture evaluation score is its \textit{premature} performance. However, this may not be the best choice of evaluation score.  For example, it has been observed that, as a network is trained, multiple sets of internal features begin to emerge over network layers, and the quality of these internal features determines the ultimate “behavioral” performance of the neural network as a whole. Based on these observations, we reasoned that, if we could evaluate the quality of a network's internal features even in a very \textit{premature} state, we might be able to more quickly determine if a given architecture is likely to obtain high levels of \textit{mature} performance. 

But without a reference set of high quality internal features, how can we determine the quality of a network's internal features?  The main idea proposed here is to use features of a high performing ``teacher" network as a reference to identify promising sample architectures at a much earlier \textit{premature} state. Our proposed method is inspired by prior work showing that the internal representations of a high-performing \textit{teacher} network can be used to optimize the parameters of smaller, shallower, or thinner “student” networks \cite{ba2014deep,hinton2015distilling,Romero2014, Carreira2018}. It is also inspired by the fact that such internal representation measures can potentially be obtained from primate brains and thus could be used as an ultimate teacher. While our ability to simultaneously record from large populations of neurons is fast growing \cite{stevenson2011advances}, these measurements have already been shown to have remarkable similarities to internal activations of CNNs \cite{Yamins2014, Schrimpf2018}. 

One challenge in comparing representations across models or between models and brains is the lack of one-to-one mapping between the features (or neurons in the brain). Representational Similarity Analysis (RSA) is a tool that summarizes the representational behavior into a matrix called Representational Dissimilarity Matrix (RDM) that embeds the distance between activation in response to different inputs. In doing so it abstracts away from individual features (i.e. activations) and therefore enables us to compare the representations originating from different models or even between models and biological organisms. 

Based on the RDM metric, we propose a method for architecture search termed ``Teacher Guided Search for Architectures by Generation and Evaluation" (TG-SAGE). Specifically, TG-SAGE guides each step of an architecture search by evaluating the similarity between representations in the candidate network and those in a fixed, high-performing teacher network with unknown architectural parameters but observable internal states.
We found that when this evaluation is combined with the usual performance evaluation (above), we can predict the “mature” performance of sampled architectures with an order of magnitude less \textit{premature} training and thus an order of magnitude less total computational cost. We then used this observation to execute multiple runs of TG-SAGE for different architecture search spaces to confirm that TG-SAGE can indeed discover network architectures of comparable mature performance to those discovered with performance-only search methods, but with far less total computational cost. More importantly, when considering the primate visual system as the teacher network with measurements of neural activity from only several hundred neural sites, TG-SAGE finds a network with an Imagenet top-1 error that was 5\% lower than that achieved by performance-guided architecture search.

In section 2 we review some of the previous studies in neural networks architecture search and the use of RSA to compare artificial and biological neural networks. In section 3 we describe representational dissimilarity matrix and how we use this metric in TG-SAGE to compare representations. In section 4 we show the effectiveness of TG-SAGE in comparison to performance-guided search methods by using two search methods in different architectural spaces of increasing size. We then show that in the absence of a teacher model, how we can use measurements from the brain as a teacher to guide the architecture search. 

\section{Previous Work}
There have been several recent studies on using reinforcement learning to design high performing neural network architectures \cite{Baker2017, Zoph2017a, Zoph2017b}. Neural Architecture Search (NAS) \cite{Zoph2017a, Zoph2017b} utilizes a long short-term memory network (LSTM) trained using REINFORCE to learn to design neural network architectures for object recognition and natural language processing tasks. 
Real et al. \cite{Real2016, Real2018} used evolutionary approaches in which samples taken from a pool of networks were engaged in a pairwise competition game. 

While most of these works have focused on discovering higher performing architectures, there has been a number of efforts emphasizing the computational efficiency in hyperparameter search. In order to reduce the computational cost of architecture search, Brock et al. \cite{Brock2017} proposed to use a hypernetwork \cite{Ha2016} to predict the layer weights for any arbitrary candidate architecture instead of retraining from random initial values. 
Hyperband \cite{hyperband} formulated hyperparameter search as a resource allocation problem and improved the efficiency by controlling the amount of resources (e.g. training) allocated to each sample. Similarly, several other methods proposed to increase the search efficiency by introducing early-stopping criteria during training \cite{baker2017practical} or by extrapolating the learning curve \cite{domhan2015}. These approaches are closely related to our proposed method in that, their main focus is to reduce the per-sample training cost. 

Several more recent methods \cite{Pham2018, Liu2018, XuanyiDong2019, Akimoto2019, Guo2018} proposed to share the trainable parameters across all candidate networks and to jointly optimize for the hyperparameters and the network weights during the search. While these approaches led to significant reduction in total search cost, they can only be applied to the spaces of network architectures in which the number of trainable weights do not change as a result of hyperparameter choices (e.g. when the number of filters in a CNN is fixed). 


On the other hand, there is a growing body of literature demonstrating the remarkable ability of deep neural networks trained on various categorization tasks in predicting neural and behavioral response patterns in primates. These models are able to predict the neural responses in parts of the primate visual \cite{Yamins2014, Cadieu2014, Cadena2017} and auditory cortices \cite{Kell2019}, explain patterns of object similarity judgements \cite{Rajalingham2015, Jozwik2017}, shape sensitivity in primates \cite{Kubilius2016}, and even to control the neural activity in a mid-level visual cortical area \cite{bashivan2019neural}. Moreover, it has been shown that the categorization performance of deep artificial neural networks is strongly correlated with their ability to predict the neural responses along the ventral visual pathway \cite{Yamins2014, arend2018single}.

Nevertheless, while these networks have largely progressed our understanding of neural computations in many parts of the primate brain, there are still significant differences between the two \cite{Goodfellow2014, Rajalingham2018}. Inspired by these observations, some recent studies have utilized brain measurements as constraints to alter the artificial neural networks behavior to be more similar to the brain \cite{Fong2018, blanchard2019}.


\begin{figure*}[t]
\centering
\includegraphics[width=\linewidth]{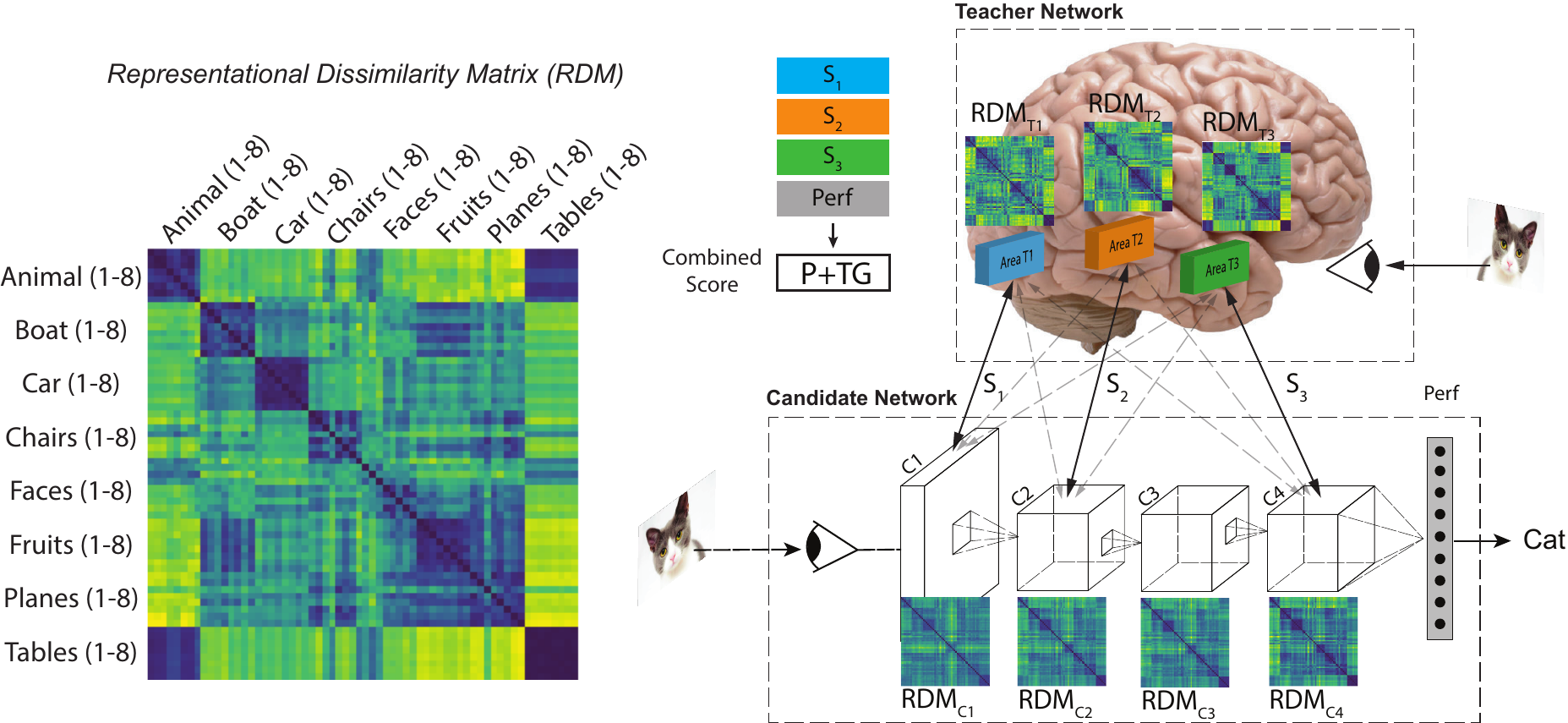}
\caption{Left -- Illustration of an exemplar RDM matrix for a dataset with 8 object categories and 8 object instances per category. Right -- Overview of TG-SAGE method. Correlation between RDMs of candidate and teacher networks, are combined with candidate network premature performance to form P+TG score for guiding the architecture search.}
\label{fig_overview}
\end{figure*}

\section{Methods}
\subsection{Representational Dissimilarity Matrix} 
\label{sec_rdm}
Representational Dissimilarity Matrix (RDM) \cite{Kriegeskorte2008} is an embedding computed for a representation that quantifies the dissimilarity between activation patterns in that representational space in response to a set of inputs or input categories.
For a given input $i$, the $n_a$ network activations at one layer can be represented as a vector $\vec{V}=f(i)\in \mathbb{R}^{n_a}$. Similarly, the collection of activations in response to a set of $n_i$ inputs can be represented in a matrix $F\in \mathbb{R}^{n_i\times n_a}$ which contains $n_a$ activations measured in response to $n_i$ inputs.

For a given activation matrix $F$, we derive RDM ($M^F$) by computing the pairwise distances between each pair of activation vectors (i.e. $F_i$ and $F_j$ which correspond to rows $i$ and $j$ in activation matrix $F$) using a distance measure like the correlation residual.

\begin{align}
\label{eq_rdm}
\centering
M^F\in \mathbb{R}^{n_i\times n_i}, M^F_{i,j}=1-corr(F_{i}, F_{j})
\end{align}

When calculating RDM for different categories (instead of individual inputs) we substitute the matrix $F$ with $F^c$ in which each row $c$ contains the average activation pattern across all inputs in category $c$.

RDM constitutes an embedding of the representational space that abstracts away from individual activations. Because of this, it allows us to compare the representations in different models or even between models and biological organisms \cite{Yamins2014, Cadieu2014}. Once RDM is calculated for two representational spaces (e.g. for a layer in each of the student and teacher networks), we can evaluate the similarity of those representations by calculating the correlation coefficient (e.g. Pearson's $r$) between the values in the upper triangle of the two RDM matrices. 

\subsection{Teacher Representational Similarity as Performance Surrogate}
\label{section_pts}
The largest portion of cost associated with neural network architecture search comes from training the sampled networks, which is proportional to the number of training steps (SGD updates) performed on the network. Due to the high cost of fully training each sampled network, in most cases a surrogate score is used as a proxy for the \textit{mature} performance. Correlation between the surrogate and \textit{mature} score may affect the architecture search performance as poor proxy values could guide the search algorithm towards suboptimal regions of the space. 
Previous work on architecture search in the space of Convolutional Neural Networks (CNN) have concurred with the empirical surrogate measure of \textit{premature} performance after about 20 epochs of training. While 20 epochs is much lower than the usual number of epochs used to fully train a CNN network (300-900 epochs), it still forces a large cost on conducing architecture searches. 
We propose that evaluating the internal representations of a network would be a more reliable measure of architecture quality during the early phase of training (e.g. after several hundreds of SGD iterations), when the features are starting to be formed but the network is not yet performing reliably on the task. 

An overview of the procedure is illustrated in Figure-\ref{fig_overview}. We evaluate each sampled candidate model by measuring the similarity between its RDMs at different layers (e.g. $M^{C1-C4}$) to those extracted from the teacher network (e.g. $M^{T1-T3}$). To this end, we compute RDM for all layers in the network and then compute the correlation between all pairs of student and teacher RDMs. 
To score a candidate network against a given layer in the teacher network, we consider the highest RDM similarity to teacher layer calculated over all layers of the student network (i.e. $S_1-S_3$; $S_i=max_j(corr(M^{Ti}, M^{Cj}))$). 

We then construct an overall teacher similarity score by taking the mean of the RDM scores which we call ``Teacher Guidance" (TG). Finally, we define the combined Performance and TG score (P+TG) which is formulated as weighted sum of \textit{premature} performance and TG score in the form of $P+\alpha TG$. 
The combined score guides the architecture search to maximize performance as well as representational similarity with the teacher architecture. The $\alpha$ parameter can be used to tune the relative weight assigned to TG score compared to the performance score. We consider the teacher architecture as any high-performing network with unknown architecture but observable activations. We can have one or several measured endpoints from the teacher network that each could potentially be used to generate a similarity score.

\section{Experiments and Results}
\subsection{Performance Predictability from Teacher Representational Similarity}
\label{section_arch_search_exp}
We first investigated if the teacher similarity evaluation measure (P+TG) of premature networks improves the prediction of \textit{mature} performance (compared to evaluation of only \textit{premature} performance, P). To do this, we made a pool of CNN architectures for which we computed the \textit{premature} and \textit{mature} performances as well as the \textit{premature} RDMs (a measure of the internal feature representation, see \ref{sec_rdm}) at every model layer.
To select the CNN architectures in the pool we first ran several performance-guided architecture searches with 20 epoch/sample training (see section \ref{sec_cnn_search} and supplementary material) and then selected 116 architectures found at different stages of the search. These networks had a wide range of \emph{mature} performance levels that also included the best network architectures found during each search.

In experiments carried out in sections 4.1 to 4.3, we used a variant of ResNet \cite{He2015} with 54 convolutional layers ($n$=9) as the teacher network. This architecture was selected as the \textit{teacher} because it is high performing (top-1 accuracy of 94.75\% and 75.89\% on CIFAR10 and CIFAR100 datasets respectively). Notably, the teacher architecture is not in our search spaces (see supp. material). Layer activations after each of the three stacks of residual blocks (here named L1-L3) were chosen as the teacher's internal feature maps. For each feature map, we took 10 random subsample of features, computed the RDM for each subsample, and then computed the average RDM across all subsamples. We did not attempt to optimize the choice of layers in the teacher network — these were chosen simply because they sampled approximately evenly over the full depth of the teacher.

\begin{figure}[hb]
\center
\includegraphics[width=\linewidth]{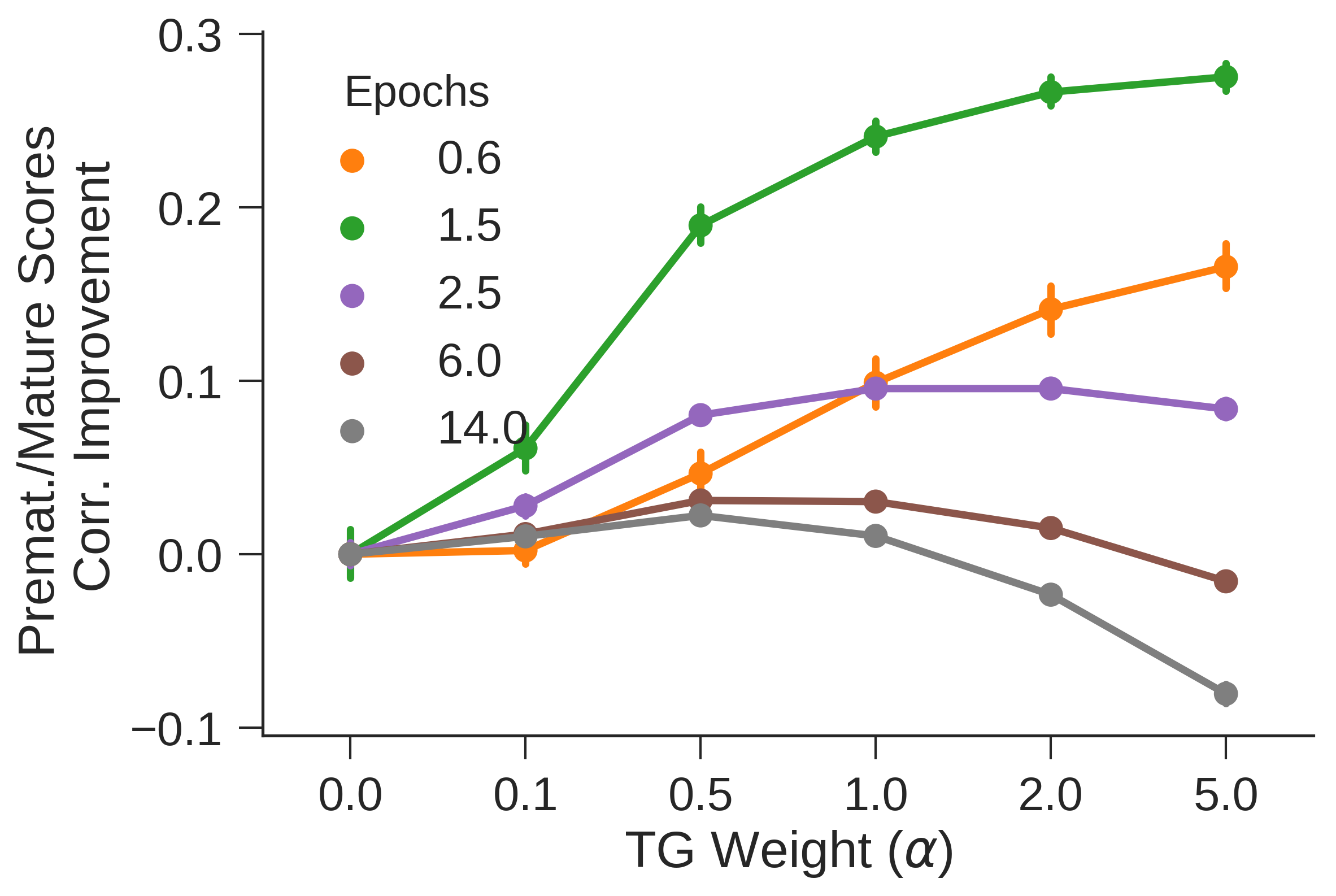}
\caption{Effect of TG weight $\alpha$ on predicting the mature performance.}
\label{fig_alpha_weights}
\end{figure}

\begin{figure}[hb]
\center
\includegraphics[width=0.45\linewidth]{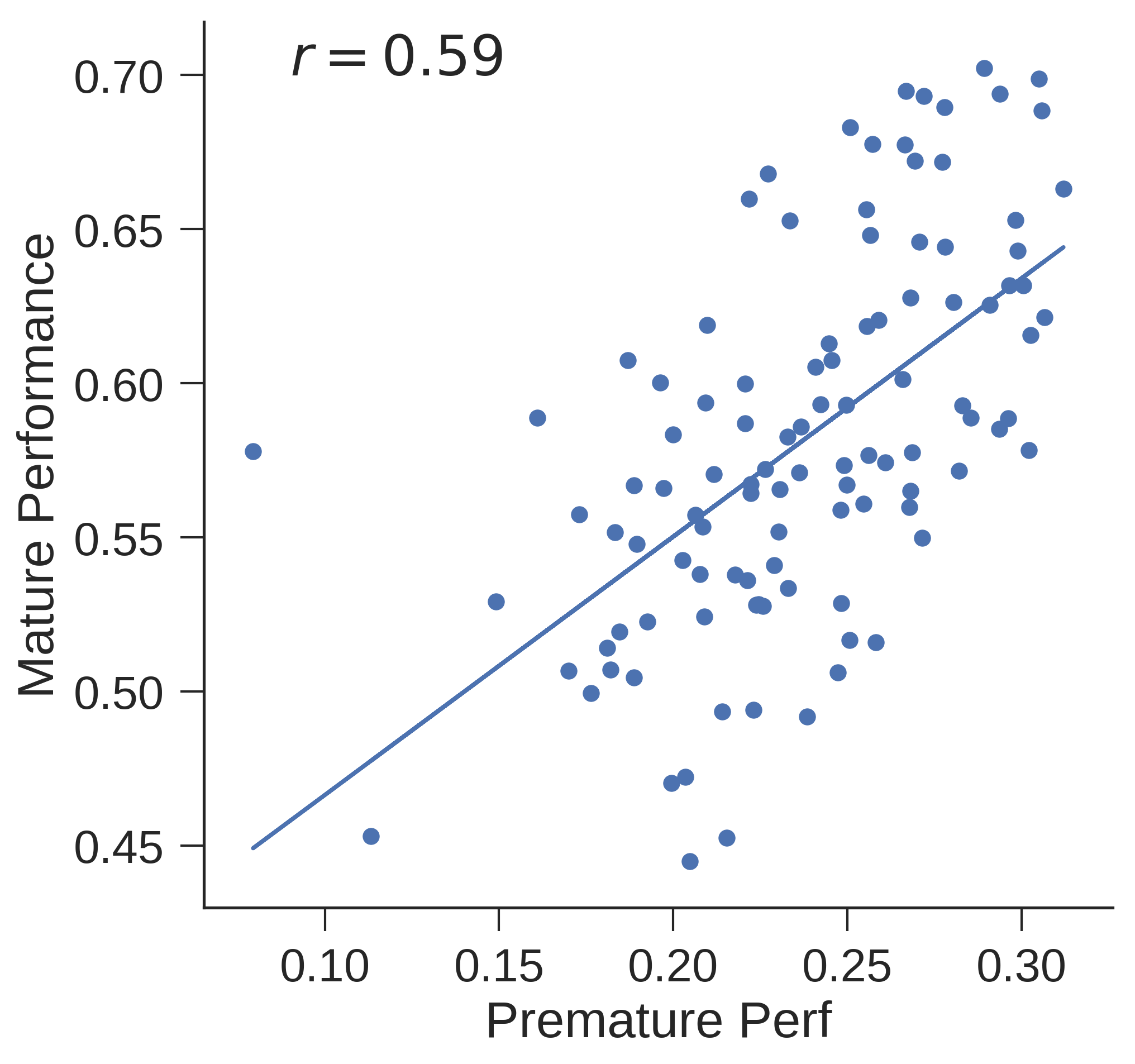}
\includegraphics[width=0.45\linewidth]{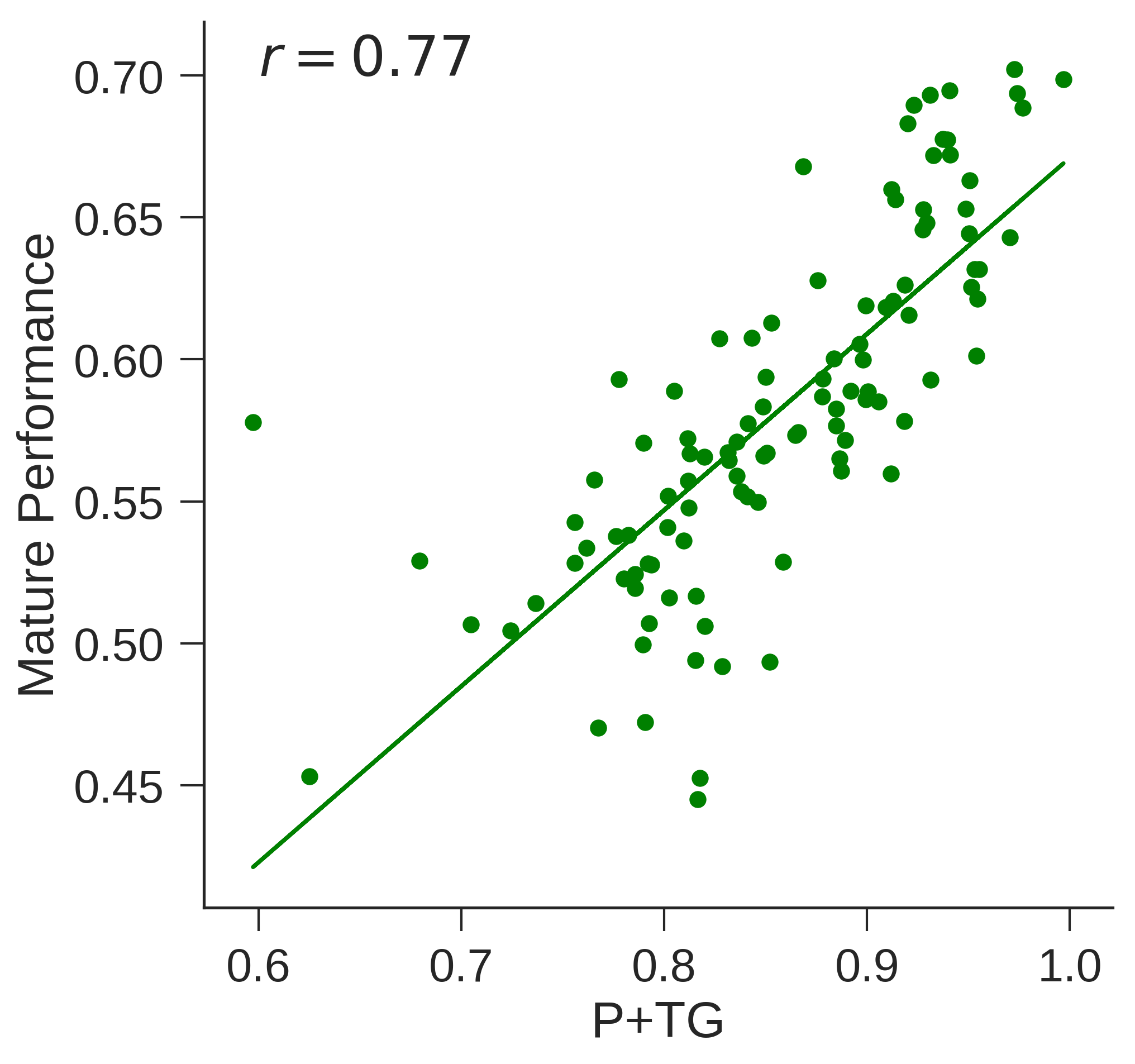}
\includegraphics[width=0.9\linewidth]{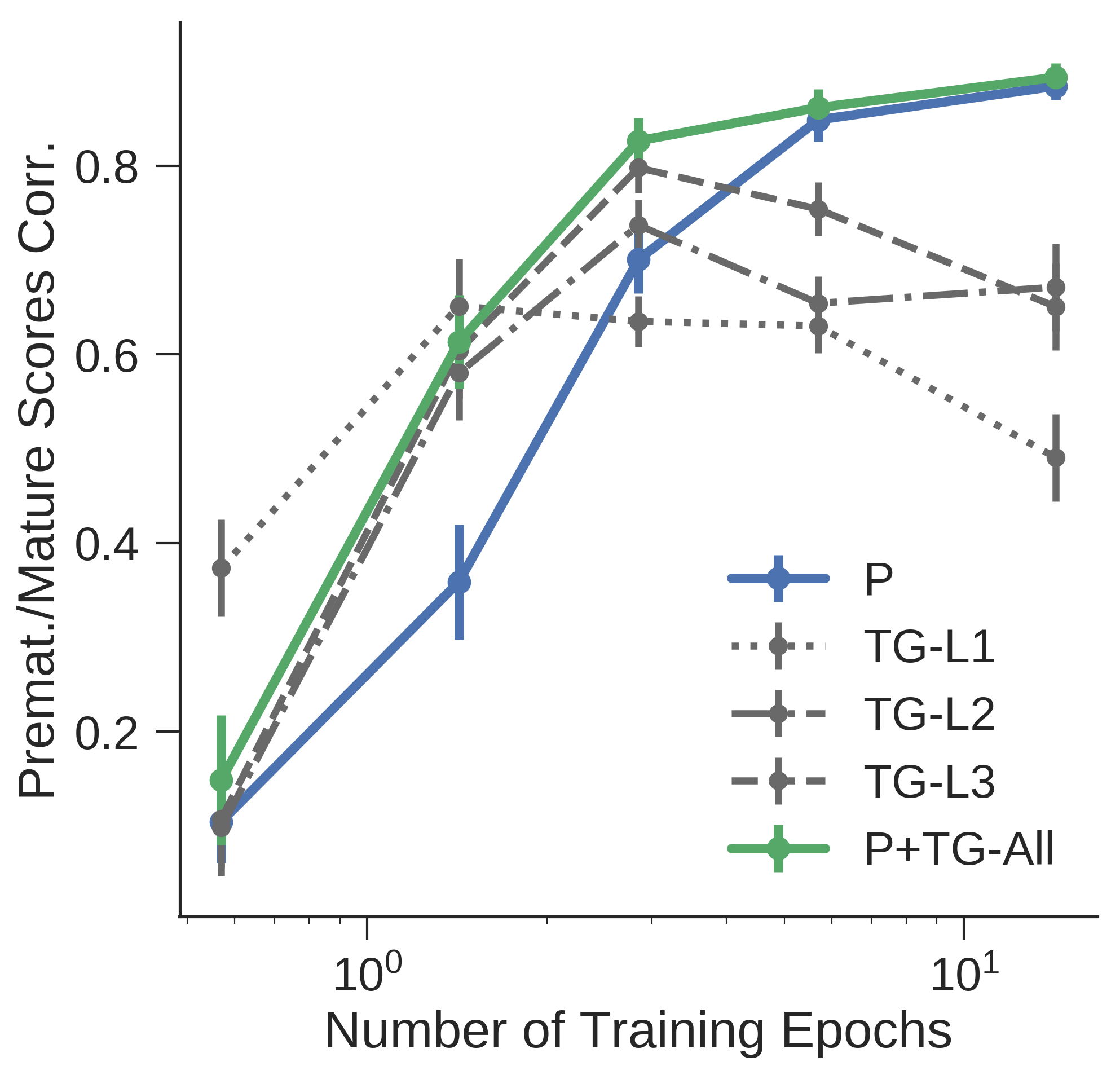}
\caption{Comparison of performance and P+TG measures at premature state (epochs=2) as predictors of mature performance. (top-left) Scatter plot of premature and mature performance values. (top-right) Scatter plot of premature P+TG measure and mature performance. (bottom) Correlation between performance, single layer RDMs, and combined P+TG measures with mature performance at varying number of premature training epochs.}
\label{fig_mat_premat_corr}
\end{figure}

In order to find the optimum TG weight factor, we varied the $\alpha$ parameter and measured the change in correlation between the P+TG score and the mature performance (Figure \ref{fig_alpha_weights}). We observed that higher $\alpha$ led to larger gains in predicting the mature performance when models were trained only for few epochs ($\le$2.5 epochs). However, with more training, larger $\alpha$ values reduced the predictability. We found that for networks trained for $\sim$2 epochs, a value of $\alpha=1$ is close to optimum. The combined ``P+TG" score (see \ref{section_pts}) composes the best predictor of \textit{mature} performance during most of the early training period (Figure \ref{fig_mat_premat_corr}-bottom). This observation was consistent with previous findings that learning in deep networks predominantly happen ``bottom-up" \cite{Raghu2017}.

We further found that the earlier teacher layers (L1) are better predictors of the \textit{mature} performance compared to other layers early on during the training ($<$2epochs) but as the training progresses, the later layers (L2 and L3) become better predictors ($\sim$3epochs) and with more training ($>$3epochs) the \textit{premature} performance becomes the best single predictor of the \textit{mature} (i.e. fully trained) performance (Figure \ref{fig_mat_premat_corr}).

In addition to ResNet, we also analyzed a second teacher network, namely NASNet (see section 2 in supp. material) and confirmed our findings using the alternative teacher network. We also found that NASNet activations (which performs higher than ResNet; 82.12\% compared to 75.9\%) form a better predictor of \textit{mature} performance in almost all training regimes (see supp. material).

\subsection{Teacher Guided Search in the Space of Convolutional Networks}
\label{sec_cnn_search}
As outlined in the Introduction, we expected that the (P+TG) evaluation score's improved predictivity (Figure \ref{fig_mat_premat_corr}) should enable it to support a more efficient architecture search than performance evaluation alone (P).  To test this directly, we used the (P+TG) evaluation score in full architectural search experiments using a range of configurations.  For these experiments, we searched two spaces of convolutional neural networks similar to previous search experiments \cite{Zoph2017a} (maximum network depth of either 10 or 20 layers). These architectural search spaces are important and interesting because they are large.  In addition, because networks in these search spaces are relatively inexpensive to train to maturity, we could evaluate the true underlying search progress at a range of checkpoints (below).  We ran searches in each space using four different search methods:  using the (P+TG) evaluation score at 2 or 20 epochs of premature training, and using the (P) evaluation score at either 2 or 20 epochs of premature training. For these experiments, we used random \cite{pinto2009high}, reinforcement learning (RL) \cite{Zoph2017a}, as well as TPE architecture selection algorithm \cite{Bergstra2011} (see Methods), and we halted the search after 1000 or 2000 sampled architectures (for the 10- and 20-layer search spaces, respectively). We conducted our search experiments on CIFAR100 instead of CIFAR10 because of larger number of classes in the dataset that provided a higher dimensional RDM. 

\begin{table*}[h]
\begin{center}
    \caption{Comparison of premature performance and representational similarity measure in architecture search using RL and TPE  algorithms. P: premature performance as validation score; P+TG: combined premature performance and RDMs as the validation score. Values are  $\mu\pm\sigma$ across 3 search runs.}
    \label{rl_tpe_comparison}
    \resizebox{0.85\textwidth}{!}{%
    \begin{tabular}{l|cccc|cc}
    \toprule
    \begin{tabular}[c]{@{}c@{}}Search Algorithm\end{tabular}     & \multicolumn{4}{c|}{\textbf{RL}}                                                & \multicolumn{2}{c}{\textbf{TPE}}                          \\ 
    \begin{tabular}[c]{@{}c@{}}Search Space\end{tabular}      & \multicolumn{2}{c}{\textbf{10 layer}} & \multicolumn{2}{c|}{\textbf{20 layer}} & \multicolumn{1}{c}{\textbf{10 layer}} & \textbf{20 layer} \\ 
    \begin{tabular}[c]{@{}c@{}} \# Epoch/Sample\end{tabular} & 2                  & 20                & 2                  & 20                & 2                                      & 2                 \\
    \midrule
    Random - Best C100 Error (\%)   & 45.4$\pm$ 2.5           & 41.3$\pm$ 1.5          & 41.2$\pm$ 1.8           & 38.3$\pm$ 4.8          &  45.4$\pm$ 2.5                    & 41.2$\pm$ 1.8          \\
    \midrule
    P - Best C100 Error (\%)        & 41.0$\pm$ 0.5           & 40.5$\pm$ 0.4          & 37.5$\pm$ 0.2           & 32.7$\pm$ 0.9          & 42.5$\pm$ 5.7                    & 37.0$\pm$ 3.0          \\
    P+TG - Best C100 Error (\%)     & \textbf{38.3$\pm$ 1.1}           & \textbf{39.2$\pm$ 0.9}          & \textbf{33.2$\pm$ 1.4}           & \textbf{32.2$\pm$ 0.8}          & \textbf{37.6$\pm$ 1.2}                    & \textbf{33.0$\pm$ 2.4 }         \\ 
    \midrule
    Performance Improvement (\%)                                                   & 2.7                & 1.3               & 4.3                & 0.5               & 4.9                                    & 4                 \\
    \bottomrule
    \end{tabular}}
\end{center}
\end{table*}

We found that, for all search configurations, the (P+TG) driven search algorithm (i.e. TG-SAGE) consistently outperformed the performance-only driven algorithm (P) in that, using equal computational cost it always discovered higher performing networks (Table \ref{rl_tpe_comparison}).  This gain was substantial in that TG-SAGE found network architectures with approximately the same performance as (P) search but at $\sim10\times$ less computational cost (2 vs. 20 epochs; Table \ref{rl_tpe_comparison}).

\begin{figure*}[h]
\centering
\includegraphics[width=2.5in]{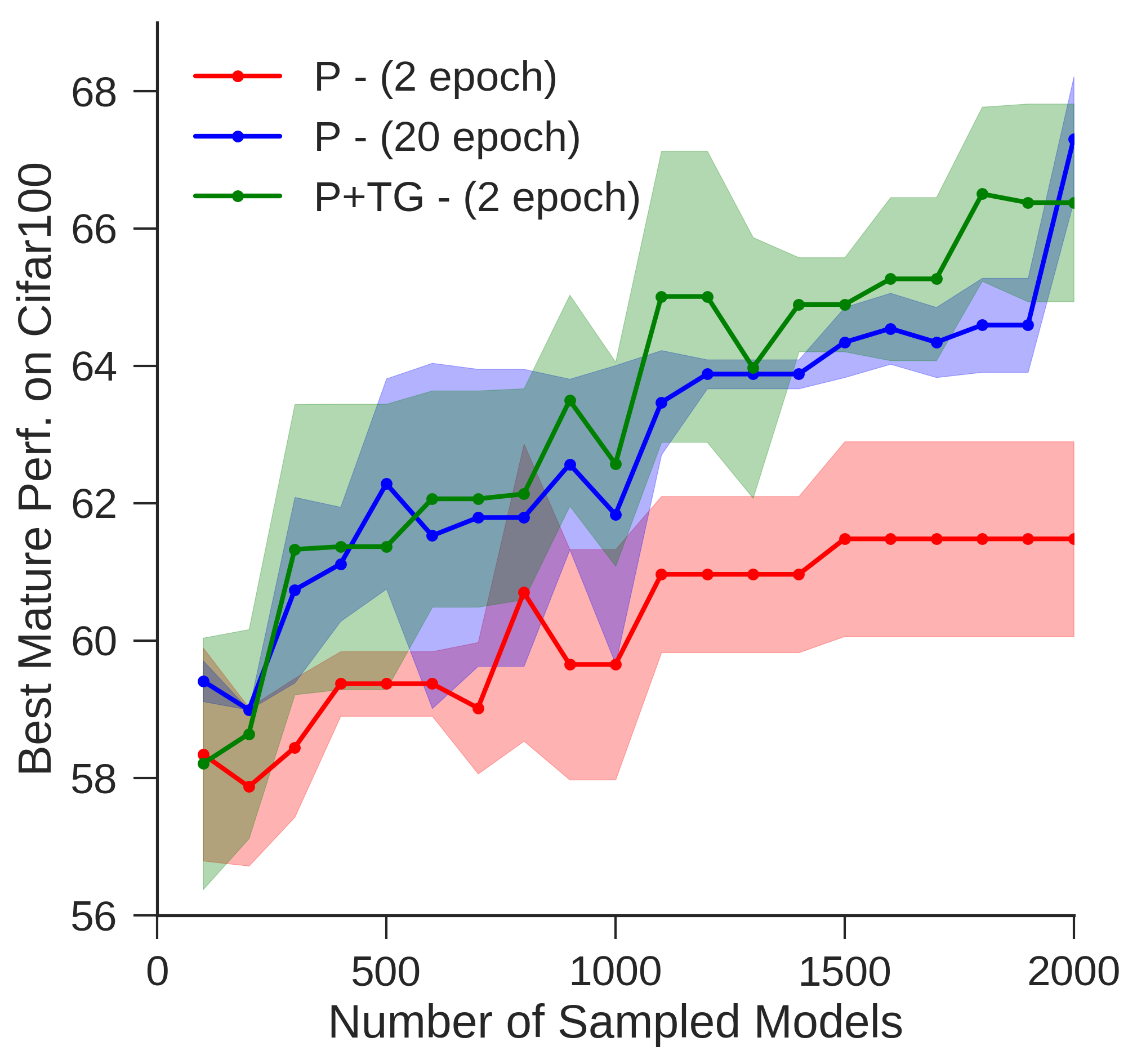}
\includegraphics[width=2.5in]{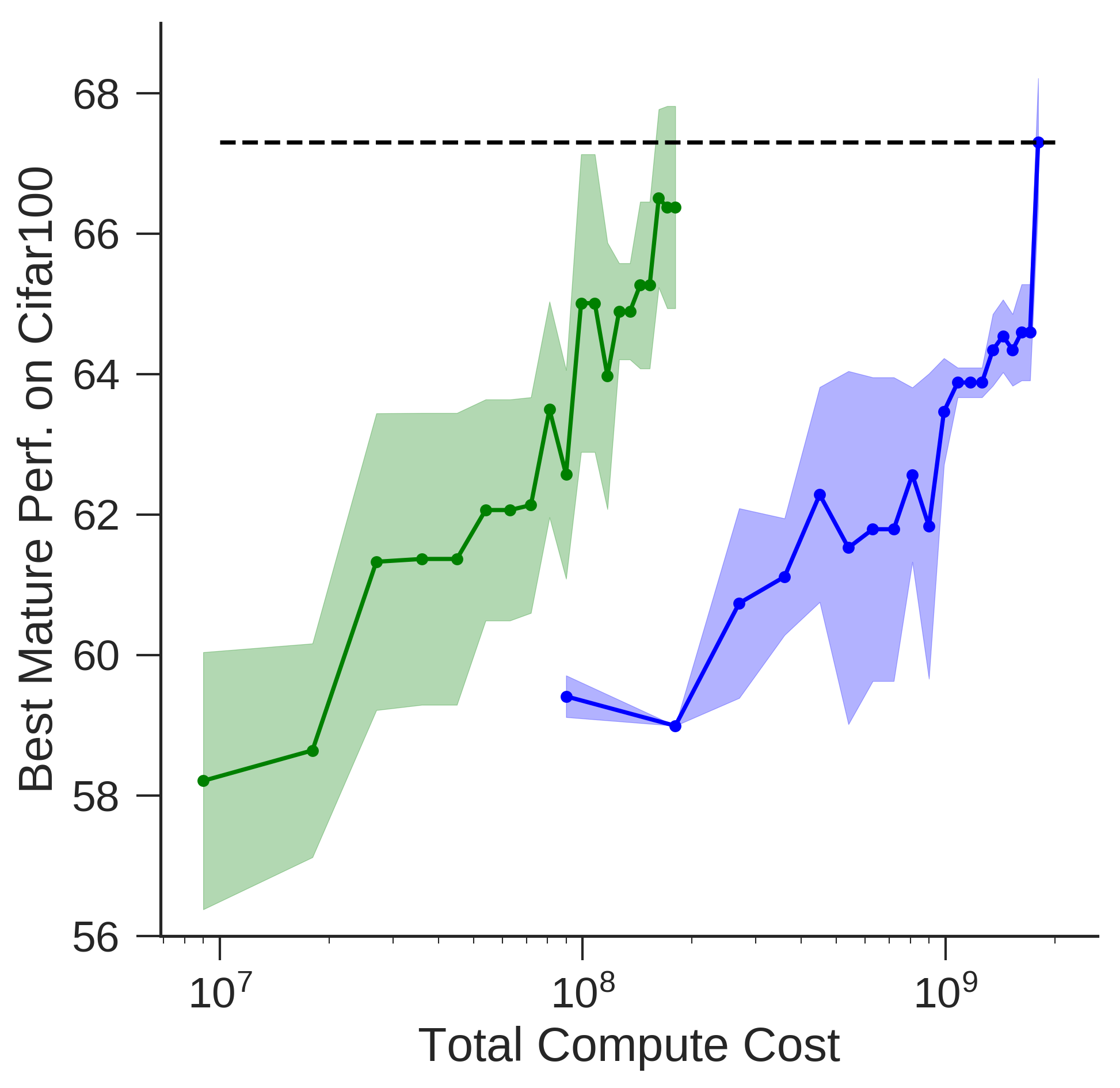}
\caption{Effect of different surrogate measures on architecture search performance. (left) shows the average C100 performance of the best network architectures found during different stages of three runs of RL search in each case (see text). (right) same as the plot on left but displayed with respect to the total computational cost invested ($\textnormal{number of training images}\times \textnormal{number of epochs} \times \textnormal{number of samples}$).}
\label{fig_search_eff}
\end{figure*}

To assess and track the efficiency of these searches, we measured the maximum validation set performance of the fully trained network architectures returned by each search at its current choice of the top-5 architectures.  
We repeated each search experiment three times to estimate variance in these measures resulting from both search sampling and network initial filter weight sampling. Figure \ref{fig_search_eff} shows that the teacher guided search (P+TG) leads to finding network architectures that were on par with performance guided search (P) throughout the search runs while being 10$\times$ more efficient. 

\subsection{Teacher Guided Search in the Space of Convolutional Cells}
\label{sec_cell_search}
To compare our approach with recent work on architecture search, we performed a search experiment with P+TG score on the space of convolutional cells \cite{Zoph2017b, Liu2017}. One advantage of this search space compared to those in section \ref{sec_cnn_search} is that convolutional cells are transferable across datasets. After a cell structure is sampled, the full architecture is constructed by stacking the same cell multiple times with a predefined structure (see supplementary material). While both RL and TPE search methods led to similar outcomes in our experiments in section \ref{section_arch_search_exp}, average TPE results were slightly higher for both experiments. Hence, we chose to conduct the search experiment in this section using TPE algorithm with the same setup as in section \ref{section_arch_search_exp} using CIFAR100 with 1000 samples.

For each sample architecture, we computed RDMs for each cell's output. Considering that we had $N=2$ cell repetitions in each block during search, we ended up with 8 RDMs in each sampled cell that were compared with 3 precomputed RDMs from the teacher network (24 comparisons over validation set of 5000 images). 
Due to the imperfect correlation between the \textit{premature} and \textit{mature} performances, doing a small post-search reranking step increases the chance of finding slightly better cell structures. We chose the top 10 discovered cells and trained them for 300 epochs on the training set (45k samples) and evaluated on the validation set (5k samples). Cell structure with the highest validation performance was then fully trained on the complete training set (50k samples) for 600 epochs similar to \cite{Zoph2017b} and evaluated on the test set. 

\begin{table*}[h]
\begin{center}
    \caption{Performance of discovered cells on CIFAR10 and CIFAR100 datasets. *indicates error rates from retraining the network using the same training pipeline on 2-GPUs. B: Number of operation blocks in each cell. N: number of cell repetitions in each network block. F: number of filters in the first cell. $^\dagger$ estimated GPU days.}
    \label{table_cifar_results}
    \resizebox{0.95\textwidth}{!}{
    \begin{tabular}{c|c|cc|cccccc}
    \toprule
    \textbf{Network}     & \textbf{\# Params}      & \textbf{C10 Error} & \textbf{C100 Error} & $\boldsymbol{M_1}$   & $\boldsymbol{E_1}$  & $\boldsymbol{M_2}$ & $\boldsymbol{E_2}$ 
    &\multicolumn{1}{c}{\begin{tabular}[c]{@{}c@{}}\textbf{Cost}\\ \textbf{(Examples)}\end{tabular}}  
    &\multicolumn{1}{c}{\begin{tabular}[c]{@{}c@{}}\textbf{Cost}\\ \textbf{(GPU days)}\end{tabular}} \\ \midrule
    AmoebaNet-A \cite{Real2018} & 3.2M       & 3.34               & -      & 20000         & 1.13M         & 100         & 27M       & 25.2B         & (813)\\ 
    NASNet-A \cite{Zoph2017b} & 3.3M       & 3.41 (3.72$^*$)               & 17.88$^*$      & 20000         & 0.9M         & 250         & 13.5M       & 21.4-29.3B       & (690)\\ 
    PNASNet-5 \cite{Liu2017} & 3.2M               & 3.41 (4.06$^*$)       & 19.26$^*$               & 1160          & 0.9M         & 0          & 0           & 1.0B          & (32)\\ 
    
    ENAS \cite{Pham2018} & 4.6M       & 3.54               & 19.43      & 310         & 50k        & 0         & 0       & 15.5M    & 0.5 \\ 
    GDAS (FRC) \cite{XuanyiDong2019} & 2.5M       & 3.75      & 19.09        & -        & -        & -        & -        & -        & 1 \\
    ASNG + cutout \cite{Akimoto2019} & 3.9M          & 2.83          & -           & -        & -        & -        & -        & -       & 0.11 \\
    DARTS + cutout \cite{Liu2018} & 3.4M       & 2.83      & -        & -        & -        & -        & -        & -   & 4 \\
    IRLAS + cutout \cite{Guo2018} & 3.4M     &2.71       & -        & -        & -        & -        & -        & -     & -\\
    \midrule
    SAGENet & 6.0M               & 3.66                & \textbf{17.42}               & \multirow{2}{*}{1000} & \multirow{2}{*}{90K} & \multirow{2}{*}{10}          & \multirow{2}{*}{13.5M}       & \multirow{2}{*}{225M}      & \multirow{2}{*}{(7)}\\ 
        SAGENet-sep  & \textbf{2.7M}               & 3.88                & 17.51               &  &  &           &        &  \\ 
    \bottomrule
    \end{tabular}}
\end{center}
\end{table*}

We compared our best found cell structure with those found using NAS \cite{Zoph2017b} and PNAS \cite{Liu2017} methods on CIFAR-10, CIFAR-100, and Imagenet datasets (Tables \ref{table_cifar_results} and \ref{table_imagenet_results}). To rule out any differences in performance that might have originated from differences in training procedure, we used the same training pipeline to train our proposed network (SAGENet) as well as the as well as the two baselines. 

With regard to compactness, SAGENet had more parameters and FLOPS compared to NASNet and PNASNet due mostly to symmetric $7\times1$ and $1\times7$ convolutions. But we had not considered any costs associated with the number of parameters or the number of FLOPS when conducting the search experiments. For this reason, we also considered another version of SAGENet in which we replaced the symmetric convolutions with ``$7\times7$ separable" convolutions (SAGENet-sep). SAGENet-sep had half the number of parameters and FLOPS compared to SAGENet and slightly higher error rates. 
To compare the cost and efficiency of different search procedures we adopted the same measures as in \cite{Liu2017}. Total cost of search was computed as the total number of examples that were processed with SGD throughout the search procedure. This includes $M_1$ sampled cell structures that were trained with $E_1$ examples during the search and $M_2$ top cells trained on $E_2$ examples post-search to find the top performing cell structure. The total cost was then calculated as $M_1E_1 + M_2E_2$.

In sum, we find that while SAGENet performed on par to both NAS and PNAS top networks on C10, C100, and Imagenet, the cost of search was about 100 and 4.5 times less than NASNet and PNASNet respectively (Table \ref{table_cifar_results}). Interestingly, at \textit{mature} state our top architecture performed better than the teacher network (ResNet) on C10 and C100 datasets (96.34\% and 82.58\% on C10 and C100 for TG-SAGE as compared to 94.75\% and 75.89\% for the ResNet). 


\begin{table*}[h]
\caption{Performance of discovered cells on Imagenet dataset in mobile settings (i.e. number of parameters $\sim 5.5 M$ and number of FLOPS$<$1.5B). Hyperparameters B, N, and F are selected so the network contains approximately 5.5M parametesrs and less than 1.5B FLOPS. *indicates error rates from training all networks using the same training pipeline on 2-GPUs. }
\begin{center}
    {%
    \resizebox{0.8\textwidth}{!}{
    \begin{tabular}{c|ccc|cc|cc}
    \toprule
    \textbf{Network}     & \textbf{B} & \textbf{N} & \textbf{F} & \textbf{\# Params (M)} & \textbf{FLOPS (B)} & \textbf{Top-1 Err$^*$} & \textbf{Top-5 Err$^*$}  \\ \midrule
    NASNet-A  & 5       & 4                 & 44       & 5.3      & 1.16      & 31.07     & 11.41      \\ 
    PNASNet-5 & 5       & 3                 & 56             & 5.4      & 1.30   &\textbf{29.92}     & \textbf{10.63}            \\ 
    SAGENet & \multirow{2}{*}{5}         & \multirow{2}{*}{4}       & \multirow{2}{*}{48}  & 9.7      & 2.15  & 31.81     & 11.79      \\ 
    SAGENet-sep &    &        &     & 4.9      & 1.03    & 31.9      & 11.99      \\ \bottomrule
    \end{tabular}}}
\end{center}
\label{table_imagenet_results}
\end{table*}

\begin{table*}[h]
\caption{Comparison of best networks found with performance-guided and neurally guided architecture searches in the space of convolutional cells on Imagenet.}
\begin{center}
    {\resizebox{0.9\textwidth}{!}{
    \begin{tabular}{c|ccc|cc|cc}
    \toprule
    \textbf{Network}     & \textbf{B} & \textbf{N} & \textbf{F} & \textbf{\# Params (M)} & \textbf{FLOPS (B)} & \textbf{Top-1 Err} & \textbf{Top-5 Err} \\ \midrule
    P-imagenet & 5   & 4       & 40  & 5.5      & 1.26 & 34.4      & 13.5       \\ 
    SAGENet-neuro & 5    & 3        & 40   & 5.6     & 1.35   & \textbf{32.54}      & \textbf{12.26}       \\ \bottomrule
    \end{tabular}}}
\end{center}
\label{table_neural_results}
\end{table*}

\subsection{Using Cortical Measurements as the Teacher Network}
In the absence of an already high performing teacher network, the utility of TG-SAGE seems unclear (why would one need this method if one already has a high-performing network implemented on a computer?). But what if one does not have that computer implementation, while has partial access to the internal activations of a high performing network?  One such network is the primate ventral visual system -- it is both high performing in object categorization tasks and is partially observable through electrophysiological recording tools. To test the utility of this idea, we conducted an additional experiment in which we used neural spiking measurements from macaque ventral visual cortex to guide the architecture search. 

To facilitate the comparison of representations between the brain's ventral stream and CNNs, we needed a fixed set of inputs that could be shown to both CNNs and the monkeys. For this purpose, we used a set of 5760 images that contained 3D rendered objects placed on uncorrelated natural backgrounds and were designed to include large variations in position, size, and pose of the objects (see supplementary material). We used previously published neural measurements from 296 neural sites in two macaque monkeys in response to these images \cite{Yamins2014, Majaj2015}. These neural responses were measured from three anatomical regions along the ventral visual pathway (V4, posterior-inferior temporal (p-IT), and anterior inferior temporal (a-IT) cortex) in each monkey -- a series of cortical regions in the primate brain that underlie visual object recognition. 
To allow the candidate networks to be more comparable to the brain measurements, we conducted the experiment on the Imagenet dataset and trained each candidate network for $1/5$ epoch using images of size $64\times 64$. We used the same setup as in section \ref{sec_cell_search} but with three RDMs generated from our neural measurements in each area (i.e. V4, p-IT, a-IT). We held out 50,000 of the images from the original Imagenet training set as the validation set that was used to evaluate the premature performance for the candidate networks. To further speed up the search, we removed the first 2 reduction cells in the architecture during the search. Similar to experiments in previous sections, we used $\alpha=1$ to weigh the RDM similarities compared to the performance in assigning scores to each candidate network. 
After running the architecture search for 1000 samples, we picked the top 10 networks and fully trained them on Imagenet for 40 epochs and picked the network with highest validation accuracy. We then trained this network on the full Imagenet training set and evaluated its performance on the test set. 
As a baseline, we also performed a similar search but using only the performance metric to guide the search. 

We found that the best discovered network using the combined P+TG metric (SAGENet-neuro) had a significantly lower top-1 error (32.54\%) than the best network derived from performance-guided search (34.4\%; P-imagenet; see Table \ref{table_neural_results}). This suggests that the approach has potential merit. However, when searching on CIFAR-100 dataset, the best model found by using the partially observed internal representations of the primate brain teacher network (SAGENet-neuro) did not perform as well as the model found by  using the fully observed internal representations of the ResNet teacher network. One critical factor that might have affected the quality of the best discovered model was the amount of per-sample training done during the search (this was limited to 2000 steps ($1/5$epoch) in our experiment). Naturally, allowing more training before evaluation would potentially result in a more accurate prediction of mature performance and discovering a higher performing model. Another important factor was the partially observations of the brain -- neural recordings for constructing a teacher RDM would likely be further improved with larger population of neural responses measured in response to more stimuli.  
Nevertheless, the teacher representations constructed from only a few hundred primate neural sites was already informative enough to produce improved guidance for architecture search.

\section{Discussion and Future Directions}
\label{discussions}
We here demonstrate that, when the internal “neural” representations of a high performing teacher neural network are partially observable (such as the brain's neural network), that knowledge can substantially accelerate the discovery of high performing artificial networks. We propose a new method to accomplish that acceleration (TG-SAGE) and demonstrate its utility using an existing state-of-the-art computerized network as the teacher (ResNet) and its potential using a partially-observed biological network as the teacher (primate ventral visual stream).  Essentially, TG-SAGE jointly maximizes a model's \textit{premature} performance and its internal representational similarity to those of a partially observable teacher network.  With the architecture space and search settings tested here, we report a computational efficiency gain of $\sim10\times$ in discovering CNNs for performance on visual categorization. This gain in search efficiency (reduced computational resource budget with similar categorization performance) was achieved without any additional constraints on the search space as in alternative search methods like ENAS \cite{Pham2018} or DARTS \cite{Liu2018}. We empirically demonstrated this by performing searches in several CNN architectural spaces. 



Could this approach be applied at scale using high performing biological systems?   We here showed how limited measurements from the brain (neural population patterns of responses to many images) could be formulated as teacher constraints to accelerate the search for higher performing networks. It remains to be seen if larger scale neural measurements -- which are obtainable in the near future -- could achieve even better acceleration.  

{\small
\bibliographystyle{ieee_fullname}
\bibliography{egbib}

\begin{thebibliography}{10}\itemsep=-1pt

\bibitem{Akimoto2019}
Youhei Akimoto, Shinichi Shirakawa, Nozomu Yoshinari, Kento Uchida, Shota
  Saito, and Kouhei Nishida.
\newblock {Adaptive Stochastic Natural Gradient Method for One-Shot Neural
  Architecture Search}.
\newblock In {\em ICML}, 2019.

\bibitem{arend2018single}
Luke Arend, Yena Han, Martin Schrimpf, Pouya Bashivan, Kohitij Kar, Tomaso
  Poggio, James~J DiCarlo, and Xavier Boix.
\newblock Single units in a deep neural network functionally correspond with
  neurons in the brain: preliminary results.
\newblock Technical report, Center for Brains, Minds and Machines (CBMM), 2018.

\bibitem{ba2014deep}
Jimmy Ba and Rich Caruana.
\newblock Do deep nets really need to be deep?
\newblock In {\em Advances in neural information processing systems}, pages
  2654--2662, 2014.

\bibitem{Baker2017}
Bowen Baker, Otkrist Gupta, Nikhil Naik, and Ramesh Raskar.
\newblock Designing neural network architectures using reinforcement learning.
\newblock {\em arXiv preprint arXiv:1611.02167}, 2016.

\bibitem{baker2017practical}
Bowen Baker, Otkrist Gupta, Ramesh Raskar, and Nikhil Naik.
\newblock Practical neural network performance prediction for early stopping.
\newblock {\em arXiv preprint arXiv:1705.10823}, 2017.

\bibitem{bardenet2010surrogating}
Remi Bardenet and Balazs Kegl.
\newblock Surrogating the surrogate: accelerating gaussian-process-based global
  optimization with a mixture cross-entropy algorithm.
\newblock In {\em 27th International Conference on Machine Learning (ICML
  2010)}, pages 55--62. Omnipress, 2010.

\bibitem{bashivan2019neural}
Pouya Bashivan, Kohitij Kar, and James~J DiCarlo.
\newblock Neural population control via deep image synthesis.
\newblock {\em Science}, 364(6439):eaav9436, 2019.

\bibitem{Bergstra2011}
James Bergstra, Remi Bardenet, Yoshua Bengio, and Balazs Kegl.
\newblock {Algorithms for Hyper-Parameter Optimization}.
\newblock pages 1--9, 2011.

\bibitem{bergstra2012machine}
James Bergstra, Nicolas Pinto, and David Cox.
\newblock Machine learning for predictive auto-tuning with boosted regression
  trees.
\newblock In {\em Innovative Parallel Computing (InPar), 2012}, pages 1--9.
  IEEE, 2012.

\bibitem{Bergstra2012}
J. Bergstra, D. Yamins, and D.~D. Cox.
\newblock {Making a Science of Model Search}.
\newblock pages 1--11, 2012.

\bibitem{bergstra2013hyperopt}
James Bergstra, Dan Yamins, and David~D Cox.
\newblock Hyperopt: A python library for optimizing the hyperparameters of
  machine learning algorithms.
\newblock In {\em Proceedings of the 12th Python in Science Conference}, pages
  13--20. Citeseer, 2013.

\bibitem{blanchard2019}
Nathaniel Blanchard, Jeffery Kinnison, Brandon RichardWebster, Pouya Bashivan,
  and Walter~J Scheirer.
\newblock A neurobiological cross-domain evaluation metric for predictive
  coding networks.
\newblock In {\em Conference on Computer Vision and Pattern Recognition}, 2019.

\bibitem{Brock2017}
Andrew Brock, Theodore Lim, J.~M. Ritchie, and Nick Weston.
\newblock {SMASH: One-Shot Model Architecture Search through HyperNetworks}.
\newblock 2017.

\bibitem{Cadena2017}
Santiago~A Cadena, George~H Denfield, Edgar~Y Walker, Leon~A Gatys, Andreas~S
  Tolias, Matthias Bethge, and Alexander~S Ecker.
\newblock {Deep convolutional models improve predictions of macaque V1
  responses to natural images Author summary}.
\newblock {\em Plos}, pages 1--28, 2017.

\bibitem{Cadieu2014}
Charles~F. Cadieu, Ha Hong, Daniel L~K Yamins, Nicolas Pinto, Diego Ardila,
  Ethan~A. Solomon, Najib~J. Majaj, and James~J. DiCarlo.
\newblock {Deep Neural Networks Rival the Representation of Primate IT Cortex
  for Core Visual Object Recognition}.
\newblock {\em PLoS Computational Biology}, 10(12), 2014.

\bibitem{Carreira2018}
Joao Carreira, Viorica Patraucean, Laurent Mazare, Andrew Zisserman, and Simon
  Osindero.
\newblock {Massively Parallel Video Networks}.
\newblock In {\em ECCV}, 2018.

\bibitem{domhan2015}
Tobias Domhan, Jost~Tobias Springenberg, and Frank Hutter.
\newblock Speeding up automatic hyperparameter optimization of deep neural
  networks by extrapolation of learning curves.
\newblock 15:3460--8, 2015.

\bibitem{Fong2018}
Ruth~C. Fong, Walter~J. Scheirer, and David~D. Cox.
\newblock {Using human brain activity to guide machine learning}.
\newblock {\em Scientific Reports}, 8(1):1--10, 2018.

\bibitem{Goodfellow2014}
Ian~J. Goodfellow, Jonathon Shlens, and Christian Szegedy.
\newblock {Explaining and Harnessing Adversarial Examples}.
\newblock pages 1--11, 2014.

\bibitem{Guo2018}
Minghao Guo, Zhao Zhong, Wei Wu, Dahua Lin, and Junjie Yan.
\newblock {IRLAS: Inverse Reinforcement Learning for Architecture Search}.
\newblock In {\em CVPR}, 2018.

\bibitem{Ha2016}
David Ha, Andrew Dai, and Quoc~V. Le.
\newblock {HyperNetworks}.
\newblock 2016.

\bibitem{He2015}
Kaiming He, Xiangyu Zhang, Shaoqing Ren, and Jian Sun.
\newblock {Deep Residual Learning for Image Recognition}.
\newblock {\em Arxiv.Org}, 7(3):171--180, 2015.

\bibitem{hinton2015distilling}
Geoffrey Hinton, Oriol Vinyals, and Jeff Dean.
\newblock Distilling the knowledge in a neural network.
\newblock {\em arXiv preprint arXiv:1503.02531}, 2015.

\bibitem{Huang2016}
Gao Huang, Zhuang Liu, Laurens van~der Maaten, and Kilian~Q. Weinberger.
\newblock {Densely Connected Convolutional Networks}.
\newblock 2016.

\bibitem{hutter2011sequential}
Frank Hutter, Holger~H Hoos, and Kevin Leyton-Brown.
\newblock Sequential model-based optimization for general algorithm
  configuration.
\newblock In {\em International Conference on Learning and Intelligent
  Optimization}, pages 507--523. Springer, 2011.

\bibitem{Jozwik2017}
Kamila~M. Jozwik, Nikolaus Kriegeskorte, Katherine~R. Storrs, and Marieke Mur.
\newblock {Deep convolutional neural networks outperform feature-based but not
  categorical models in explaining object similarity judgments}.
\newblock {\em Frontiers in Psychology}, 8(OCT):1726, 2017.

\bibitem{Kell2019}
Alexander~JE Kell and Josh~H. McDermott.
\newblock {Deep neural network models of sensory systems: windows onto the role
  of task constraints}.
\newblock {\em Current Opinion in Neurobiology}, 55:121--132, 2019.

\bibitem{Kingma2014}
Diederik Kingma and Jimmy Ba.
\newblock {Adam: A Method for Stochastic Optimization}.
\newblock {\em International Conference on Learning Representations}, pages
  1--13, 2014.

\bibitem{Kriegeskorte2008}
Nikolaus Kriegeskorte, Marieke Mur, and Peter~a. Bandettini.
\newblock {Representational similarity analysis - connecting the branches of
  systems neuroscience.}
\newblock {\em Frontiers in systems neuroscience}, 2(November), 2008.

\bibitem{Krizhevsky2012}
Alex Krizhevsky, Ilya Sutskever, and Geoffrey~E Hinton.
\newblock {ImageNet Classification with Deep Convolutional Neural Networks}.
\newblock {\em Advances In Neural Information Processing Systems}, pages 1--9,
  2012.

\bibitem{Kubilius2016}
Jonas Kubilius, Stefania Bracci, and Hans~P. {Op de Beeck}.
\newblock {Deep Neural Networks as a Computational Model for Human Shape
  Sensitivity}.
\newblock {\em PLoS Computational Biology}, 12(4):1--26, 2016.

\bibitem{hyperband}
Lisha Li, Kevin Jamieson, Giulia DeSalvo, Afshin Rostamizadeh, and Ameet
  Talwalkar.
\newblock Hyperband: A novel bandit-based approach to hyperparameter
  optimization.
\newblock {\em The Journal of Machine Learning Research}, 18(1):6765--6816,
  2017.

\bibitem{Liu2017}
Chenxi Liu, Barret Zoph, Jonathon Shlens, Wei Hua, Li-Jia Li, Li Fei-Fei, Alan
  Yuille, Jonathan Huang, and Kevin Murphy.
\newblock {Progressive Neural Architecture Search}.
\newblock 2017.

\bibitem{Liu2018}
Hanxiao Liu, Karen Simonyan, and Yiming Yang.
\newblock {DARTS: Differentiable Architecture Search}.
\newblock 2018.

\bibitem{Majaj2015}
N.~J. Majaj, H. Hong, E.~A. Solomon, and J.~J. DiCarlo.
\newblock {Simple Learned Weighted Sums of Inferior Temporal Neuronal Firing
  Rates Accurately Predict Human Core Object Recognition Performance}.
\newblock {\em Journal of Neuroscience}, 35(39):13402--13418, 2015.

\bibitem{Pascanu2012}
Razvan Pascanu, Tomas Mikolov, and Yoshua Bengio.
\newblock {On the difficulty of training Recurrent Neural Networks}.
\newblock (2), 2012.

\bibitem{Pham2018}
Hieu Pham, Melody~Y. Guan, Barret Zoph, Quoc~V. Le, and Jeff Dean.
\newblock {Efficient Neural Architecture Search via Parameters Sharing}.
\newblock 2018.

\bibitem{pinto2009high}
Nicolas Pinto, David Doukhan, James~J DiCarlo, and David~D Cox.
\newblock A high-throughput screening approach to discovering good forms of
  biologically inspired visual representation.
\newblock {\em PLoS computational biology}, 5(11):e1000579, 2009.

\bibitem{Raghu2017}
Maithra Raghu, Justin Gilmer, Jason Yosinski, and Jascha Sohl-Dickstein.
\newblock {SVCCA: Singular Vector Canonical Correlation Analysis for Deep
  Learning Dynamics and Interpretability}.
\newblock 2017.

\bibitem{Rajalingham2018}
Rishi Rajalingham, Elias~B Issa, Pouya Bashivan, Kohitij Kar, Kailyn Schmidt,
  and James~J Dicarlo.
\newblock {Large-scale, high-resolution comparison of the core visual object
  recognition behavior of humans, monkeys, and state-of-the-art deep artificial
  neural networks}.
\newblock {\em The Journal of neuroscience}, 014970(33):240614, 2018.

\bibitem{Rajalingham2015}
R. Rajalingham, K. Schmidt, and J.~J. DiCarlo.
\newblock {Comparison of Object Recognition Behavior in Human and Monkey}.
\newblock {\em Journal of Neuroscience}, 35(35):12127--12136, 2015.

\bibitem{Real2018}
Esteban Real, Alok Aggarwal, Yanping Huang, and Quoc~V Le.
\newblock {Regularized Evolution for Image Classifier Architecture Search}.
\newblock (2017), 2018.

\bibitem{Real2016}
Esteban Real, Sherry Moore, Andrew Selle, Saurabh Saxena, Yutaka~Leon Suematsu,
  Quoc Le, and Alex Kurakin.
\newblock {Large-Scale Evolution of Image Classifiers}.
\newblock 2016.

\bibitem{Romero2014}
Adriana Romero, Nicolas Ballas, Samira~Ebrahimi Kahou, Antoine Chassang, Carlo
  Gatta, and Yoshua Bengio.
\newblock {FitNets: Hints for Thin Deep Nets}.
\newblock pages 1--13, 2014.

\bibitem{Schrimpf2018}
Martin Schrimpf, Jonas Kubilius, Ha Hong, Najib~J Majaj, Rishi Rajalingham,
  Elias~B Issa, and Kohitij Kar.
\newblock {Brain-Score : Which Artificial Neural Network for Object Recognition
  is most Brain-Like ?}
\newblock pages 1--9, 2018.

\bibitem{Simonyan2014}
Karen Simonyan and Andrew Zisserman.
\newblock {Very Deep Convolutional Networks for Large-Scale Image Recognition}.
\newblock pages 1--10, 2014.

\bibitem{stevenson2011advances}
Ian~H Stevenson and Konrad~P Kording.
\newblock How advances in neural recording affect data analysis.
\newblock {\em Nature neuroscience}, 14(2):139, 2011.

\bibitem{Szegedy2014}
Christian Szegedy, Vincent Vanhoucke, Jonathon Shlens, and Zbigniew Wojna.
\newblock {Rethinking the Inception Architecture for Computer Vision}.
\newblock 2014.

\bibitem{williams1992simple}
Ronald~J Williams.
\newblock Simple statistical gradient-following algorithms for connectionist
  reinforcement learning.
\newblock {\em Machine learning}, 8(3-4):229--256, 1992.

\bibitem{XuanyiDong2019}
Yi~Yang {Xuanyi Dong}.
\newblock {Searching for A Robust Neural Architecture in Four GPU Hours |
  Xuanyi Dong}.
\newblock {\em Computer Vision and Pattern Recognition 2019}, pages 1761--1770,
  2019.

\bibitem{Yamins2014}
D.~L.~K. Yamins, H. Hong, C.~F. Cadieu, E.~A. Solomon, D. Seibert, and J.~J.
  DiCarlo.
\newblock {Performance-optimized hierarchical models predict neural responses
  in higher visual cortex}.
\newblock {\em Proceedings of the National Academy of Sciences},
  111(23):8619--8624, 2014.

\bibitem{Zoph2017a}
Barret Zoph and Quoc~V Le.
\newblock {Neural architecture Search With reinforcement learning}.
\newblock {\em ICLR}, 2017.

\bibitem{Zoph2017b}
Barret Zoph, Vijay Vasudevan, Jonathon Shlens, and Quoc~V. Le.
\newblock {Learning Transferable Architectures for Scalable Image Recognition}.
\newblock 10, 2017.

\end{thebibliography}
}

\clearpage

\renewcommand\thefigure{S\arabic{figure}}  
\renewcommand\thetable{S\arabic{table}}  
\setcounter{figure}{0}
\section*{Supplementary Material}
\section*{Hyperparameter Search with Reinforcement Learning (RL)}
\label{NAS}
We follow the method proposed by \cite{Zoph2017a} to learn the probability of hyperparameter choices ($\mathcal{X}={x_1, x_2, ..., x_n}$) that maximize the unknown but observable reward function $f: \mathcal{X} \rightarrow\mathbb{R}$. A 2-layer long short-term memory (LSTM) is used as the controller that chooses each hyperparameter in the network at every unrolling step. The LSTM network, models the conditional probability distribution of optimal hyperparameter choices as a function of all previous choices $P(x_j\vert x_1,x_2, ...,x_{j-1}, \theta)$ in which $\theta$ is the set of all tunable parameters in the LSTM network. Since a differentiable loss function is not known for this problem, usual maximum likelihood methods could not be used in this setting. Instead parameters are optimized through reinforcement learning based approaches (e.g. REINFORCE \cite{williams1992simple}) by increasing the likelihood of each hyperparameter choice according to the reward (score) computed for each sampled network (or a batch of sampled networks). Relative to \cite{Zoph2017a}, we made two modifications. First, since the order of dependencies between different hyperparameters in each layer/block is arbitrary, we ran the LSTM controller for one step per layer (instead of once per hyper-parameter). This results in shorter choice sequences generated by the LSTM controller and therefore shorter sequence dependencies. Second, we chose a Boltzman policy method for action selection to allow the search to continue the exploration throughout the search experiment. Hyperparameter values were selected according to the probability distribution over all action choices. Compared to $\epsilon$-greedy method, following the softmax policy reduces the likelihood of sub-optimal actions throughout the training. 

For each hyperparameter, choice probability is computed using a linear transformation (e.g. $W_{K_h}, W_{N_{filters}}$) from LSTM output at the last layer ($h_l^2$) followed by a softmax. To reduce the number of tunable parameters and more generalization across layers, we used shared parameters between layers.
\begin{align}
\label{eq_choice_prob}
&\hat{P}_{l,x} = \textnormal{softmax}(W_{t}^Th_l^2) \\
&\nonumber l\in\{1,2,...,N_l\} \\
&\nonumber t\in\{K_h, K_w, N_{filters}, \textnormal{stride, normalization, activation}\}
\end{align}

Probability distribution over possible number of layers is formulated as a function of the first output value of the LSTM ($\hat{P}_{N_l} = \textnormal{softmax}(W_{N_l}^Th_0^2)$). In addition to layers' hyperparameters we also search over layers' connections. Similar to the approach taken in \cite{Zoph2017a} we formulated the probability of a connection between layer \textit{i} and \textit{j} as a function of the state of the LSTM at each of these layers ($h_i^2, h_j^2$).  
\begin{equation}
\label{eq_conn_prob}
\hat{P}^c_{i,j} = \textnormal{sigmoid}(W_{src}^Th_i^2 + W_{dst}^Th_j^2)
\end{equation}
where $\hat{P}^c_{i,j}$ represents the probability of a connection between layer \textit{i} output to \textit{j}'s input. $W_{src}$ and $W_{dst}$ are tunable parameters that link the hidden state of LSTM to probability of a connection existing between the two layers. 

\section*{Hyperparameter Search with Tree of Parzen Estimators (TPE)}
\label{section_tpe}
Sequential Model-Based Optimization (SMBO) \cite{hutter2011sequential} approaches are numerical methods used to optimize a given score function $f: \mathcal{X}\rightarrow\mathbb{R}$. They are usually applied in settings where evaluating the function at each point is costly and it's important to minimize the number of evaluations to reach the optimal value. Various SMBO approaches were previously proposed \cite{bergstra2012machine, bardenet2010surrogating} and some have been used for hyperparameter optimization in neural networks \cite{Bergstra2011, Bergstra2012, Liu2017}. 
Bayesian SMBO approaches model the posterior or conditional probability distribution of values (scores) and use a criteria to iteratively suggest new samples while the probability distribution is updated to incorporate the history of previous sample tuples $(x, y)$ where $x=(x^{(1)}, ..., x^{(n)})$ is a sample hyperparameter vector and $y$ is the received score (or loss).
Here we adopted Tree of Parzen Estimators (TPE) because of its intuitiveness and successful application in various domains with high dimensional spaces. Unlike most other Bayesian SMBO methods that directly model the posterior distribution of values $P(y\vert x)$, TPE models the conditional distribution $P(x\vert y)$ with two non-parametric densities. 
\begin{equation}
P(x\vert y) = \begin{cases}
				l(x) \qquad \forall \ y \le y^*\\
               g(x) \qquad \forall \ y > y^*\\
            \end{cases}
\end{equation}
We consider $y$ to be the loss value which we are trying to minimize (e.g. error rate of a network on a given task). For simplicity, value of $y^*$ could be taken as some quantile of values observed so far ($\gamma$). At every iteration, TPE fits a kernel density estimator with Gaussian kernels to subset of observed samples with lowest loss value ($l(x)$) and another to those with highest loss ($g(x)$). Ideally we want to find $x$ that minimizes $y$. 
Expected Improvement (EI) is the expected reduction in $f(x)$ compared to threshold $y^*$ under current model of $f$. Maximizing EI, encourages the model to further explore parts of the space which lead to lower loss values and can be used to suggest new hyperparameter samples. 
\begin{align}
\label{ei_eq1}
\nonumber EI(x) = \int^{y^*}_{-\infty }{(y^*-y)P(y\vert x)dy} \\
= \frac{ \int^{y^*}_{-\infty }{(y^*-y)P(x\vert y)P(y)}dy}{P(x)}
\end{align}

Given that $P(y<y^*)=\gamma$ and $P(x\vert y)=l(x)$ for $y<y^*$, it has been shown \cite{Bergstra2011} that EI would be proportional to $\Big(\gamma+\frac{g(x)}{l(x)}(1-\gamma)\Big)^{-1}$. Therefore the EI criterion can be maximized by taking samples with minimum probability under $g(x)$  and maximum probability under $l(x)$. For simplicity, at every iteration $n_{d}$ samples are drawn from $l(x)$ and the hyperparameter choice with lowest $g(x)/l(x)$ ratio is suggested as the next sample.

\section*{Alternative Teacher Network - NASNet}
\begin{figure}[ht]
\center
\includegraphics[width=0.8\linewidth]{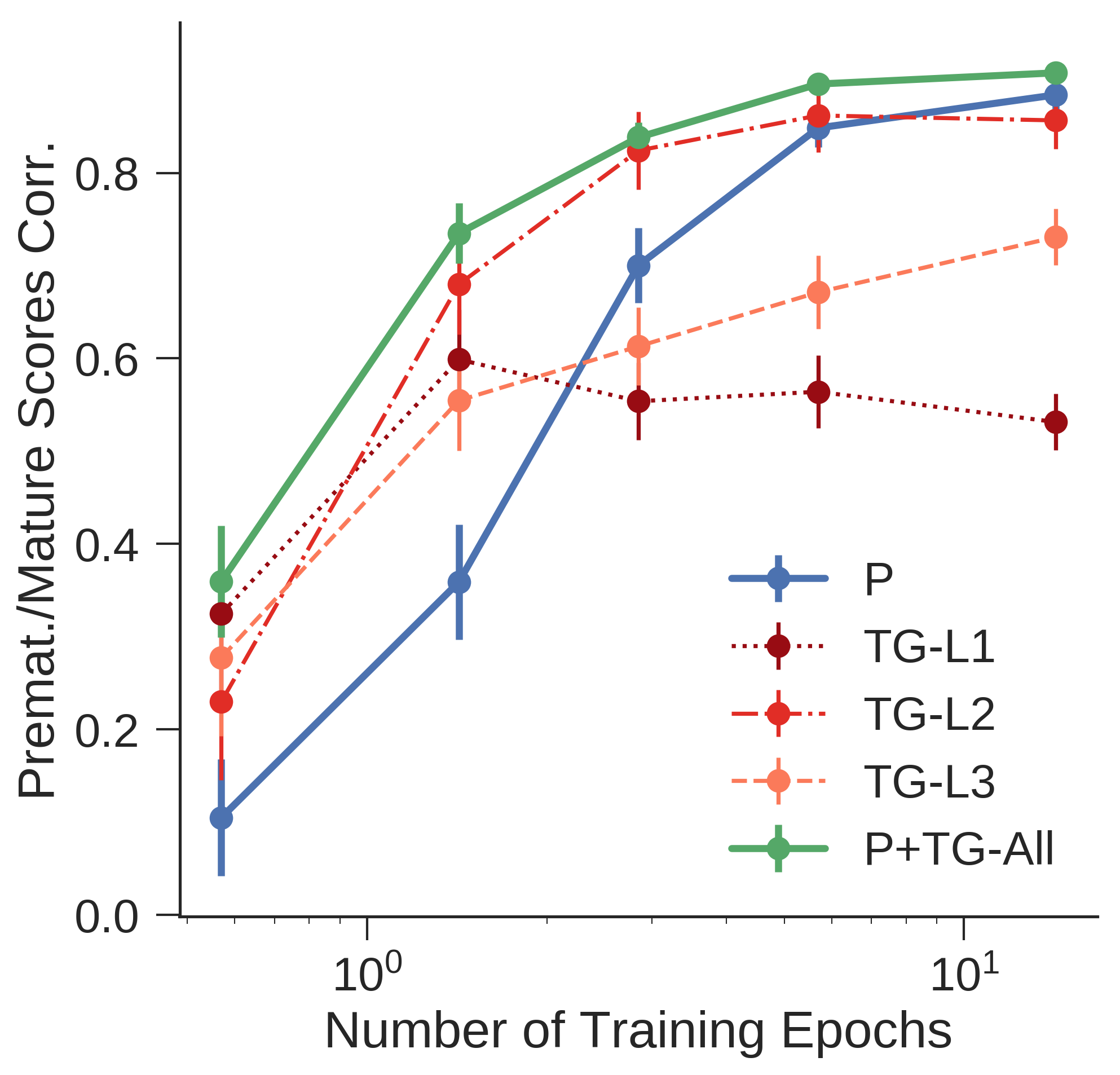}
\includegraphics[width=0.85\linewidth]{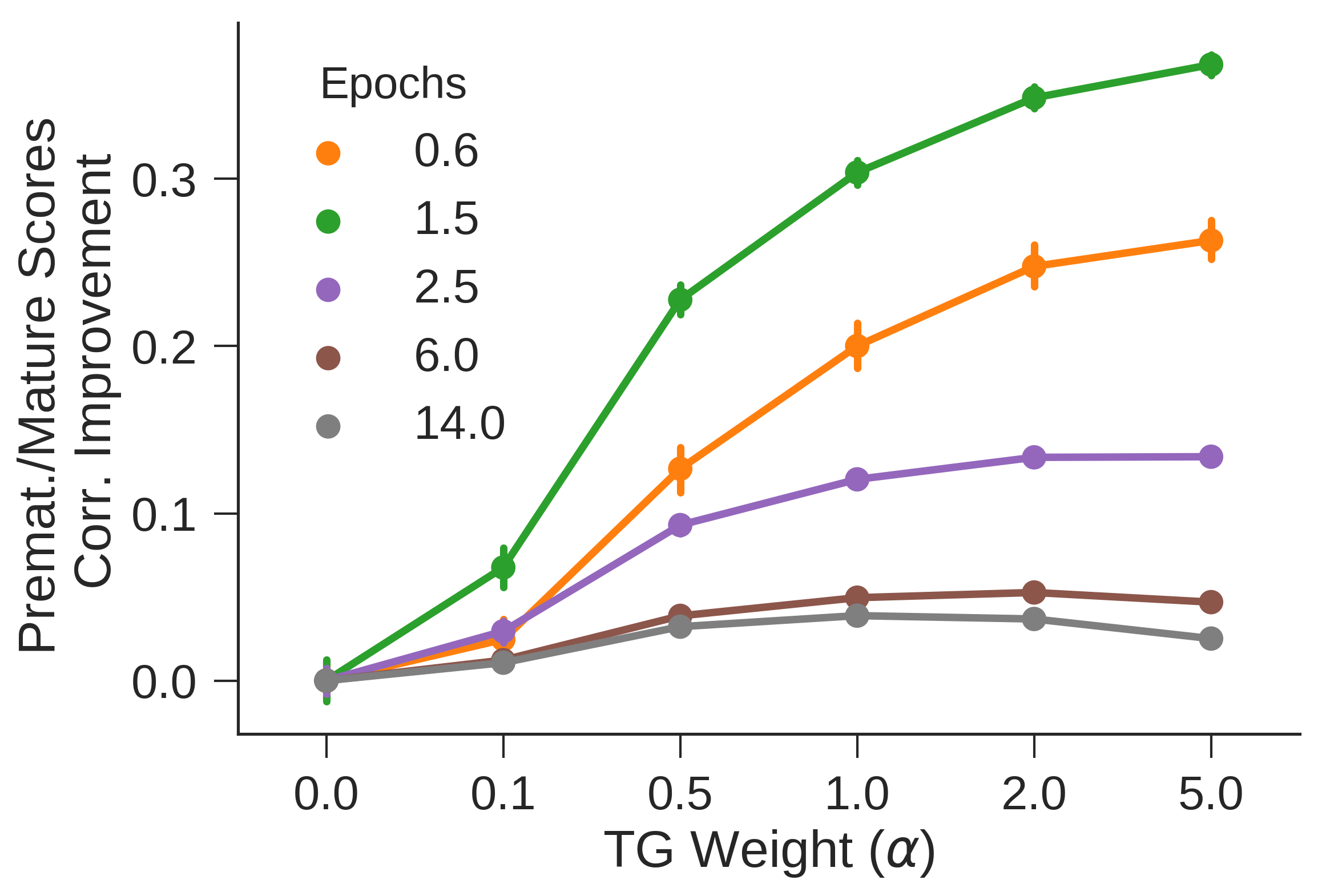}
\includegraphics[width=0.8\linewidth]{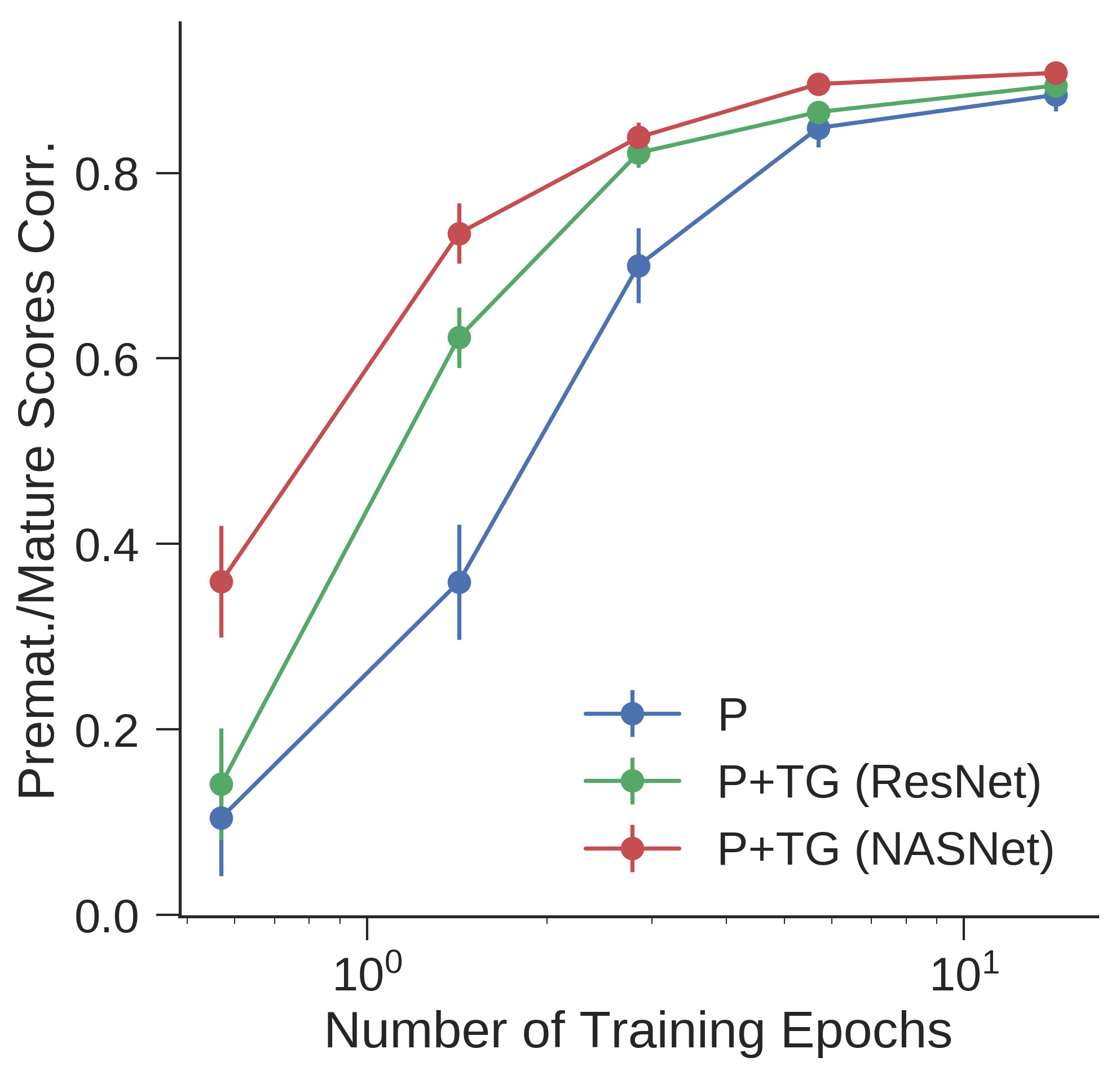}
\caption{(top) Comparison of single layer and combined RDMs with premature performance as predictors of mature performance on NASNet. P+TG was computed using $\alpha=5$.
(middle) Gain in predicting the mature performance with varying TG weight.
(bottom) Comparison of combined RDM scores using two alternative teacher models at various stages of training. $\alpha$ values of 1 and 5 were used for ResNet and NASNet respectively.}
\label{fig_weights_analysis}
\end{figure}

We examined the effect of choosing an alternative teacher network, namely NASNet and performed a set of analyses similar to those done on ResNet. We observed that similar to ResNet, early layers are better predictors of the mature performance during early stages of the training. With additional training, the premature performance becomes a better single-predictor of the mature performance but during most of the training the combined P+TG score best predicts the mature performance (Figure \ref{fig_weights_analysis}-left). We also varied the ``TG" weight factor and found that compared to ResNet, higher $\alpha$ values led to increased gains in predicting the mature performance. $\alpha=5$ was used to compute the P+TG scores shown in Figure \ref{fig_weights_analysis}. 

Overall, we found that NASNet representations were significantly better predictors of mature performance for all evaluated time points during training when compared to ResNet (Figure \ref{fig_weights_analysis}-right). 

\section*{Datasets and Preprocessing}
\textbf{CIFAR:} We followed the standard image preprocessing for CIFAR labeled dataset, a 100-way object classification task \cite{He2015}. Images were zero-padded to size $40\times40$. A random crop of size $32\times32$ was selected, randomly flipped along the vertical axis, and standardized over all pixel values in each image to have zero mean and standard deviation of 1. We split the training set into training set (45,000 images) and a validation set (5,000 images) by random selection.  

\textbf{Imagenet:} We used standard VGG preprocessing \cite{Simonyan2014} on images from Imagenet training set. During training, images were resized to have their smaller side match a random number between 256 and 512 while preserving the aspect ratio. A random crop of size 224 was then cut out from the image and randomly flipped along the central vertical axis. The central crop of size 224 was used for evaluation.

\section*{Details of Search Algorithms}
\textbf{RL Search Algorithm:} We used a two-layer LSTM with 32 hidden units in each layer as the controller. Parameters were trained using Adam optimizer \cite{Kingma2014} with a batch size of 5. For all searches, the learning rate was 0.001, and the Adam first momentum coefficient was zero $\beta_1=0$. Gradients were clipped according to global gradient norm with a clipping value of 1 \cite{Pascanu2012}.

\textbf{TPE Search Algorithm:} We used the python implementation of TPE hyperparameter search from HyperOpt package \cite{bergstra2013hyperopt}. We employed the linear sample forgetting as suggested in \cite{Bergstra2012} and set the threshold $y^*=\sqrt{N}/4$ for the set of $N$ observed samples. Each search run started with 20 random samples and continued with TPE suggestion algorithm. At every iteration, $n_d=24$ draws were taken from $l(x)$ and choice of hyperparameter $argmin_i{g(x_i)/l(x_i)}$ was used as the next sample (see section 3.3 in the main text).

\section*{Experimental Details for Search in the Space of Convolutional Networks}
\textbf{Search Space:} Similar to \cite{Zoph2017a} we defined the hyperparameter space as the following independent choices for each layer: $N_{filters}\in[32, 64, 128]$, $(K_{width}, K_{height})\in[1,3,5,7], K_{stride}\in[1,2], \textnormal{activation}\in[Identity,ReLU], \textnormal{normalization}\in[none, BN]$. In addition we searched over number of layers ($N_{layers}\in[1, N_{L}]$) and possible connections between the layers. In this space of CNNs, the input to every layer could have originated from the input image or the output of any of the previous layers. We considered two particular spaces in our experiments that differed in the value of $N_L$ (=10 or 20).

\textbf{CIFAR Training:}  Selected networks were trained on CIFAR training set (45k samples) from random initial weights using SGD with Nesterov momentum of 0.9 for 300 epochs on the training set. The initial learning rate was 0.1 and was divided by 10 after every 100 epochs. Mature performance was then evaluated on the validation set (above).

\section*{Experimental Details for Search in the Space of Convolutional Cells}
\textbf{Search Space:}
We used the same search space and network generation procedure as in \cite{Zoph2017b, Liu2017} with the exception that we added two extra hyperparameters which could force each of the cell inputs (from previous cell or the one prior to that) to be directly concatenated in the output of the cell even if they were already connected to some of the blocks in the cell. This extra hyperparameter choice was motivated by the open-source implementation of NASNet at the time of conducting the search experiments that contained similar connections\footnote{available at \url{https://github.com/tensorflow/models/blob/376dc8dd0999e6333514bcb8a6beef2b5b1bb8da/research/slim/nets/nasnet/nasnet_utils.py}}.

Each cell receives two inputs which are the outputs of the previous two cells. In early layers, the missing inputs are substituted by the input image. Each cell consists of $B$ blocks with a prespecified structure. Each block receives two inputs, an operation is applied on each input independently and the results are added together to form the output of the block. The search algorithm picks each of the operations and inputs for every block in the cell. Operations are selected from a pool of 8 possible choices: \{identity, $3\times3$ average pooling, $3\times3$ max pooling, $3\times3$ dilated convolution, $1\times7$ followed by $7\times1$ convolution, $3\times3$ depthwise-separable convolution, $5\times5$ depthwise-separable convolution, $7\times7$ depthwise-separable convolution\}.

\textbf{Imagenet Training:}
For our Imagenet training experiments, we used a batch size of 128 images of size $224\times224$ pixels. Each batch was divided between two GPUs and the gradients computed on each half were averaged before updating the weights. We used an initial learning rate of 0.1 with a decay of 0.1 after every 15 epochs. Each network was trained for 40 epochs on the Imagenet training set and validated on the central crop for all images from Imagenet validation. No dropout or drop-path was used when training the networks. RMSProp optimizer with a decay rate of 0.9 and momentum rate of 0.9 was used during training and gradients were normalized by their global norm when the norm value exceeded a threshold of 10. L2-norm regularizer was applied on all trainable weights with a weight decay rate of $4\times10^{-5}$.

\textbf{CIFAR Training:}
The networks were trained on CIFAR10/CIFAR100 training set including all 50,000 samples for 600 epochs with an initial learning rate of 0.025 and a single period cosine decay \cite{Zoph2017b}. We used SGD with Nesterov momentum rate of 0.9. We used L2 weight decay on all trainable weights with a rate of $5\times10^{-4}$. Gradient clipping similar to that used for Imagenet and a threshold of 5 was used.

\begin{figure}[th]
\center
\includegraphics[width=1\linewidth]{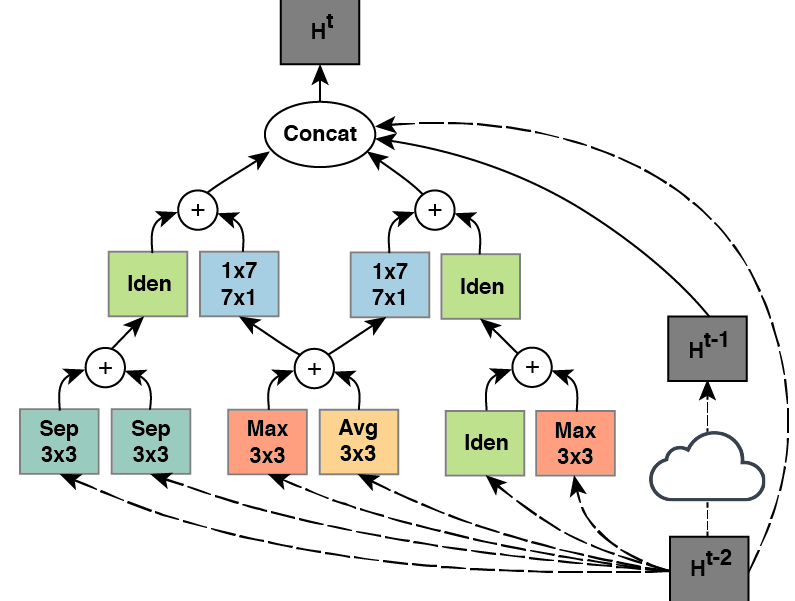}
\caption{SAGENet - Structure of the best cell discovered during the search with TG-SAGE. }
\label{fig_cell_struct}
\end{figure}

\textbf{Best Discovered Convolutional Cell:}
Figure \ref{fig_cell_struct} shows the structure of the best discovered cell by TG-SAGE on CIFAR100. Only four (out of ten) operations contain trainable weights and there are several bypass connections in the cell. 

\section*{Neural Measurements from Macaque Monkeys}
We used a dataset of neural spiking activity for a population of 296 neural sites in two awake behaving macaque monkeys in response to 5760 images \cite{Yamins2014}. Neural data were collected using parallel microelectrode arrays that were chronically implanted on the cortical surface in area V4 and IT. Fixating animals were presented with images for 100ms, and the neural response patterns were obtained by averaging the spike counts in the time window of 70-170ms post stimulus onset. To enhance the signal-to-noise ratio, each image was presented to each monkey between 21-50 times and the average response pattern across all presentation were considered for each image. The 296 recorded sites were partitioned into three cortical regions (V4, posterior-IT, and anterior-IT) and a RDM was calculated for each region. 

\begin{figure}[h]
\begin{center}
\includegraphics[width=1\linewidth]{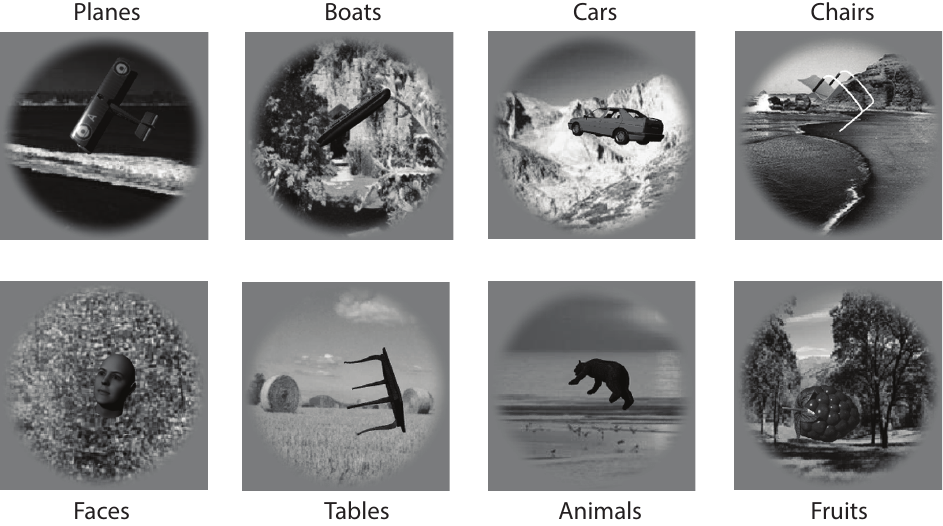}
\end{center}
\caption{Example images from each of the eight object categories that were used to record neural responses.}
\label{figure_hvm_images}
\end{figure}

The image set consisted of a total of 5760 images. Each image contained a 3D rendered object placed on an uncorrelated natural background. The rendered objects were selected from a battery of 64 objects from 8 categories (animals, boats, cars, chairs, faces, fruits, planes, and tables) with 8 objects per category. The images were generated to include large variations in position, size, and pose of the objects and were shown within the central 8$^\circ$ of monkeys' visual field. Some example images are illustrated in Figure-\ref{figure_hvm_images}. 

\section*{Implementation Details}
\label{implementation}
Because of heavy computational load associated with training neural networks and in particular in large-scale model training, we needed a scalable and efficient framework to facilitate the search procedure. We implemented our proposed framework in four main modules: (i) explorer, (ii) trainer, (iii) evaluator, and (iv) tracker. The explorer module contained the search algorithm. The trainer module optimized the parameters of the proposed architecture on an object recognition task using a large-scale image dataset. Once the training job was completed, the evaluator module extracted the network activations in response to a set of predetermined image-set and assessed the similarity of representations to the teacher benchmarks. The tracker module consisted of a database which tracked the details and status of every proposed architectures and acted as a bridge between all three modules. 

\begin{figure}[h]
\begin{center}
\includegraphics[width=1\linewidth]{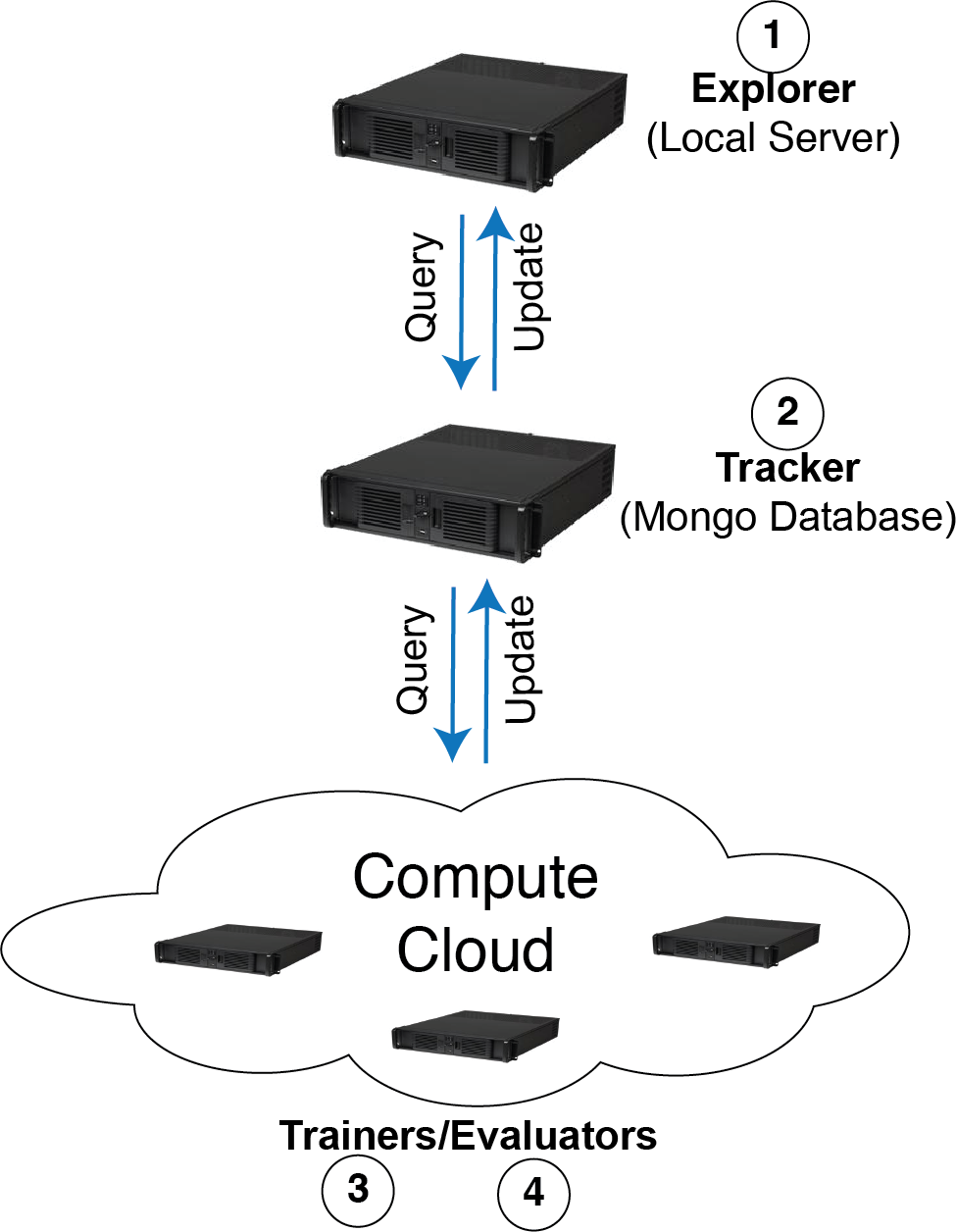}
\end{center}
\caption{Implementation of a distributed framework for conducting architecture search.}
\label{figure_servers}
\end{figure}

During the search experiments, the explorer module proposes new candidate architectures and records the details in the database (tracker module). It also continuously monitors the database for newly evaluated networks. Upon receiving adequate number of samples (i.e. when a new batch is complete), it updates its parameters. Active workers periodically monitor the database for newly added untrained models, and train the architecture on the prespecified dataset. After the training phase is completed, the evaluator module extracts the features from all layers in response to the validation set and computes the premature-performance and RDM consistencies and writes back the results in the database. The trainer and evaluator modules are then freed up to process new candidate networks. This framework enabled us to run many worker programs on several clusters speeding up the search procedure. An overview of the implemented framework is illustrated in Figure \ref{figure_servers}. Experiments reported in this paper were run on three server clusters with up to 40 GPUs in total.

\end{document}